\documentclass[a4paper,fleqn]{cas-dc}

\usepackage[authoryear]{natbib}
\usepackage{booktabs}
\usepackage{tabularx}
\usepackage{array}
\usepackage{graphicx}
\usepackage{multirow}
\usepackage{makecell}
\usepackage{algorithm}
\usepackage{algpseudocode}
\usepackage{enumitem}
\usepackage{mathtools}
\usepackage{lastpage}

\AtBeginDocument{%
}

\def\tsc#1{\csdef{#1}{\textsc{\lowercase{#1}}\xspace}}
\tsc{WGM}
\tsc{QE}
\tsc{EP}
\tsc{PMS}
\tsc{BEC}
\tsc{DE}

\begin{document}
\let\WriteBookmarks\relax
\def\floatpagepagefraction{1}
\def\textpagefraction{.001}

\shorttitle{Towards Proactive Detection of User-Side Implicit Conflicts in Human--LLM Dialogue}

\shortauthors{Wang et~al.}

\title [mode = title]{Towards Proactive Detection of User-Side Implicit Conflicts in Human--LLM Dialogue}



%
\author[1]{Jinqiang Wang}[orcid=0000-0002-3711-0743]

\fnmark[1]

\ead{jqwang@xs.ustb.edu.cn}

\credit{Conceptualization, Methodology, Investigation, Writing
– original draft}

\affiliation[1]{organization={School of Computer \& Communication Engineering, University of Science and Technology Beijing},
    city={Beijing},
    postcode={100083}, 
    country={China}}

\author[2]{Tao Zhu}[orcid=0000-0002-5879-5980] \ead{tzhu@usc.edu.cn}
\credit{Validation, Supervision}

\affiliation[2]{organization={School of Computer Science, University of South China},
    city={Hengyang},
    postcode={421001}, 
    country={China}}

\author[1]{Huansheng Ning}[orcid=0000-0001-6413-193X]
\cormark[1]
\ead{ninghuansheng@ustb.edu.cn}
\credit{Conceptualization, Supervision}

\cortext[cor1]{Corresponding author}



\begin{abstract}
In human--LLM dialogue, follow-up user utterances may implicitly conflict with earlier intents, leading the LLM to misinterpret user needs and generate inappropriate responses. A reliable dialogue system should proactively detect user-side conflicts before generating a response and seek clarification when necessary. However, prior work has largely focused on LLM-side conflicts, leaving user-side conflicts underexplored. To fill this gap, we construct UC-Bench, a human-annotated benchmark for evaluating user-side conflict detection. Preliminary experiments show that existing LLMs struggle with this task, especially when conflicts arise from implicit incompatibilities grounded in dialogue history. To improve lightweight LLMs with limited training data, we investigate data synthesis for user-side conflict detection. Existing synthesis methods do not explicitly model the implicit incompatibilities between historical and current user utterances, making it difficult to capture the evolution of conflicts and to generate reliably labeled implicit conflict samples. We propose SynUC, a constraint-guided synthesis method that represents user-side conflicts in a constraint space and uses the SPEAKING framework to guide traceable constraint transformations. Applying SynUC to WildChat, we construct UC-Data, a user-side conflict training set containing 2,487 samples. On UC-Bench, Qwen3.5-4B trained on UC-Data outperforms larger general-purpose LLMs such as Claude Opus 4.8, as well as the same backbone trained on data synthesized by existing methods.
\end{abstract}




\maketitle

\section{Introduction}

Large language models (LLMs) are increasingly used for code generation \citep{seo2026papercode} and document writing \citep{mysore-etal-2025-prototypical}. However, most existing systems still follow a passive interaction paradigm \citep{wu2026excuse, wang2025llm, zhang2026individual, li2025twostage} and respond to explicit requests without assessing whether newly introduced requirements are compatible with those established earlier in the dialogue. In human--LLM interactions, users often specify their needs incrementally by supplementing, revising, or shifting their intents across turns \citep{li2025structflowbench}. Without proactively identifying incompatibilities introduced during this process, an LLM may simply follow the latest utterance and generate responses that deviate from the user's underlying intent.
\par
Recent studies have begun to examine instruction conflicts in human--LLM dialogue. ConInstruct \citep{coninstruct} focuses on conflicts among constraints within a single user utterance, while MultiTurnInstruct \citep{multiturninstruct} evaluates models' abilities to retrieve information, track dialogue states, and resolve conflicts in multi-turn entangled instructions. However, these studies mainly assess instruction-following behavior or response quality, leaving implicit conflict relations in user-side contexts underexplored. In this work, implicit conflict refers to a situation in which a user's new requirement is incompatible with earlier requirements, but the user does not explicitly indicate a revision.
\par
\begin{figure}
    \centering
    \includegraphics[width=1\linewidth]{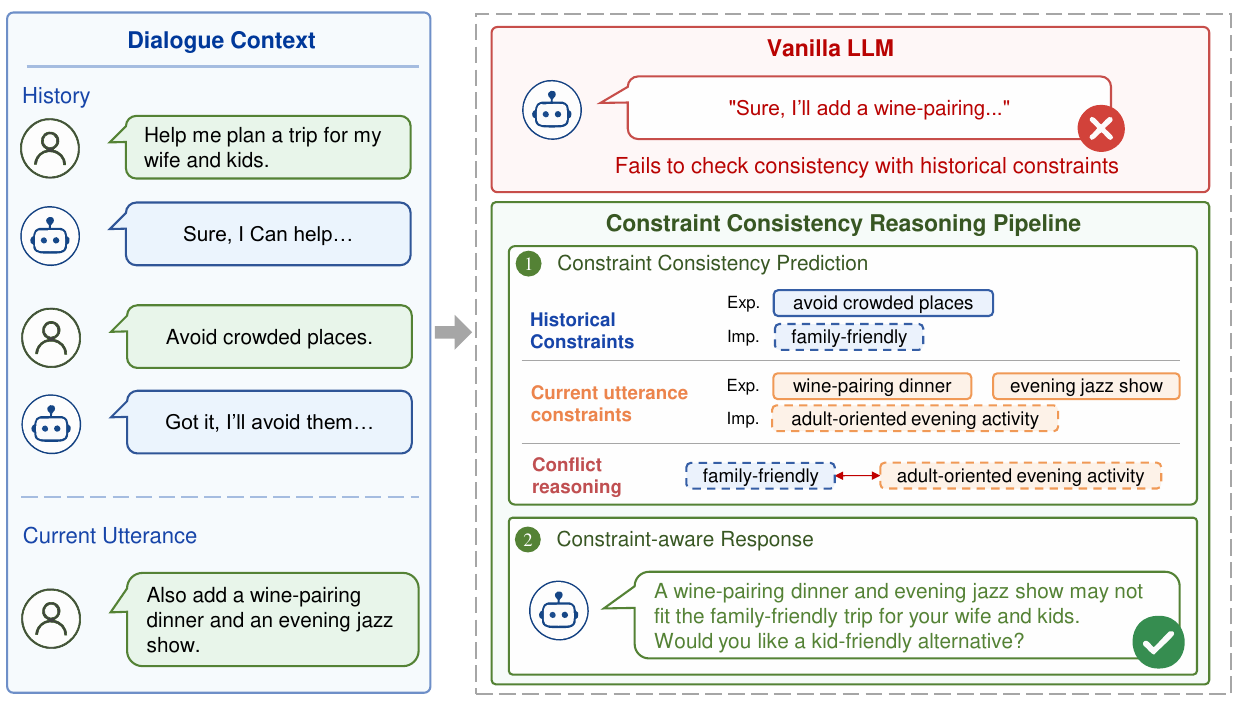}
    \caption{Comparison between a vanilla LLM and a constraint reasoning pipeline when handling an utterance containing an implicit conflict.}
    \label{fig:example}
\end{figure}

As shown in Figure \ref{fig:example}, the user first asks the LLM to “plan a trip for my wife and kids” and further specifies that it should “avoid crowded places.” This establishes an explicit constraint to avoid crowded locations and an implicit constraint that the itinerary should be family-friendly. However, the follow-up utterance requests a “wine-pairing dinner” and a “late-night jazz show.” Although the utterance does not explicitly revise the target participants, the requested activities may conflict with the earlier family-friendly goal. Detecting this conflict allows the LLM to request clarification rather than directly incorporating activities that overlook the family-friendly constraint.
\par
To evaluate the ability of LLMs to detect user-side implicit conflicts, we first construct \textbf{UC-Bench}, a human-annotated benchmark. UC-Bench categorizes the relation between the current user utterance and the dialogue history as normal dialogue, explicit revision, or implicit conflict. Normal dialogue indicates that the current utterance is compatible with previous requirements, whereas explicit revision involves an explicit modification, replacement, or cancellation of those requirements.
\par
Based on UC-Bench, we conduct a preliminary study. The results show that LLMs often fail to recognize the underlying incompatibility when directly responding to user utterances containing implicit conflicts. In contrast, their detection performance improves when they are explicitly prompted to examine potential conflicts between the current utterance and the dialogue history. These findings motivate us to formulate user-side conflict detection as a standalone task performed by an external monitoring model, which assesses this relation before response generation.
\par
\begin{figure}
    \centering
    \includegraphics[width=1\linewidth]{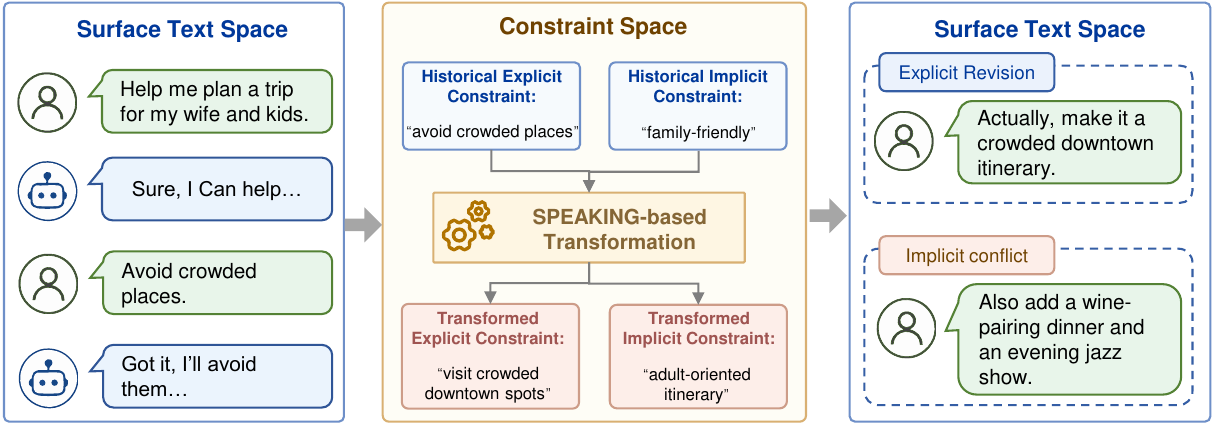}
    \caption{Illustration of the constraint-guided synthesis idea of SynUC.}
    \label{fig:idea}
\end{figure}
To address the scarcity of training data for user-side conflict detection, we investigate the automatic synthesis of high-quality training samples. Existing data synthesis methods \citep{multiturninstruct,coninstruct,hellwig2025exploring} mainly focus on general instruction generation, task response generation, or explicit conflict construction. They lack structured modeling of implicit incompatibilities between the dialogue history and the current user utterance, making it difficult to capture conflict evolution and generate controllable implicit conflict samples. To address this limitation, we propose \textbf{SynUC}, a constraint-guided data synthesis method for user-side conflict detection. As shown in Figure \ref{fig:idea}, SynUC first maps user instructions from \textbf{surface text space} to \textbf{constraint space}, where it performs constraint transformation to derive conflict constraints. It then maps these constraints back to surface text space to form conflict-bearing follow-up user utterances. The constraint transformation module is built on Hymes's SPEAKING framework \citep{SPEAKING} and modeled as a cross-sample self-evolving memory, allowing high-quality constraint conflict construction experience to be reused. Based on WildChat, we use SynUC to construct \textbf{UC-Data}, a training set with 2,487 samples for supervised fine-tuning and reinforcement learning. To further improve reinforcement learning on this task, we introduce a constraint-attribution reward that encourages the model to identify the conflicting constraints.
\par
Experimental results show that Qwen3.5-4B, a lightweight model trained on SynUC-synthesized data, outperforms a range of strong baselines on UC-Bench, including the larger closed-source model Claude Opus 4.8. Compared with other data synthesis methods, SynUC yields training data that more effectively improves implicit conflict recognition. Further ablation studies show that both constraint-space modeling and the SPEAKING-based memory contribute to the final performance, supporting the effectiveness of our method for user-side conflict detection.
\par
The main contributions of this paper are as follows.
\begin{enumerate}
    \item We present the first study of user-side implicit conflicts in multi-turn human--LLM dialogue. We construct UC-Bench, a benchmark for user-side conflict detection, and show that existing LLMs still face substantial challenges on this task.
    \item We propose SynUC, a constraint-guided data synthesis method. On UC-Bench, lightweight LLMs trained on SynUC-synthesized data outperform larger general-purpose LLMs and the same backbone trained on data synthesized by existing methods.
    \item We release UC-Data, UC-Bench, and the SynUC synthesis code to support future research. \footnote{The code and data are available at \url{https://github.com/jqwangai/SynUC}}
\end{enumerate}

The remainder of this paper is organized as follows. Section 2 reviews related work. Section 3 presents the task definition, key terminology, and preliminary study. Section 4 introduces the SynUC data synthesis method and the constraint-attribution reward. Section 5 describes the experimental setup. Section 6 reports the main results and analysis. Section 7 discusses key design choices and parameter effects. Section 8 concludes the paper.

\section{Related Work}
\subsection{Dialogue Conflict Detection}
Prior work on dialogue conflict detection has largely examined whether model responses are consistent with persona information, dialogue history, or external knowledge. In open-domain dialogue, this problem is often formulated as natural language inference. Dialogue NLI \citep{welleck2019dialogue} frames persona consistency as a natural language inference task and classifies sentence pairs derived from persona descriptions and dialogue utterances as entailment, neutral, or contradiction. DECODE \citep{nie2021like} detects whether the final utterance contradicts the preceding dialogue context. CDConv \citep{zheng2022cdconv} introduces a Chinese multi-turn dialogue benchmark covering three contradiction types: intra-sentence contradiction, role confusion, and history contradiction. In task-oriented dialogue, CI-ToD \citep{qin2021don} evaluates whether model responses are inconsistent with the user query, dialogue history, or knowledge base. Wang et al. \citep{wang2025improving} further explore zero-shot LLM prompting and multi-agent collaboration for this task.
\par
More recent studies investigate how LLMs handle contradictions in generated content and external knowledge. Mündler et al. \citep{ndler2024selfcontradictory} analyze self-contradiction in instruction-tuned language models under open-ended generation and question-answering settings, and propose prompt-based methods for detection and mitigation. CIDER \citep{CIDER} provides inconsistent dialogue responses, natural-language explanations, and clarification-based recovery utterances to support the detection and resolution of conversational inconsistencies. Wen et al. \citep{wen2024red} propose contradictory dialogue processing, in which models detect and explain self-contradictions and then modify the contradictory content. WikiContradict \citep{hou2024wikicontradict} evaluates whether LLMs can generate answers that accurately reflect conflicts among retrieved Wikipedia passages. From a data-generation perspective, ConsistentChat \citep{chen2025consistentchat} uses human conversational intents and structured information flows to guide the synthesis of consistent multi-turn instruction data. In instruction following, ConInstruct \citep{coninstruct} evaluates LLMs' ability to detect and resolve explicit conflicts among constraints within a single user instruction.
\par
Overall, existing studies primarily address inconsistencies in model responses and explicit conflicts within single-turn instructions. However, they do not specifically model implicit user-side conflicts in multi-turn interactions, where a follow-up utterance conflicts with still-valid requirements established earlier in the dialogue without an explicit revision signal.

\subsection{Data Synthesis}
Recent work has widely used LLMs to synthesize instruction data for improving instruction following. Self-Instruct \citep{wang2023self} bootstraps instructions, inputs, and outputs from a model itself, and uses the filtered data for instruction tuning. Baize \citep{xu2023baize} adopts a self-chat strategy, where ChatGPT plays both the user and the assistant to generate multi-turn dialogue data. UltraChat \citep{ding2023enhancing} further expands the scale and topic coverage of synthetic dialogue data through a systematic framework for generating large-scale, multi-topic, multi-turn instructional conversations. MAGPIE \citep{xu2025magpie} shows that an aligned LLM can autoregressively generate user instructions from only a dialogue template prefix and then produce corresponding responses, enabling large-scale extraction of instruction-response data from aligned models.
\par
Beyond scaling data, recent studies have increasingly emphasized quality control and generation logic. Evol-Instruct \citep{xu2024wizardlm} evolves simple instructions into more complex and diverse ones through iterative rewriting, improving models’ ability to follow complex instructions. REFED \citep{REFED} extracts transferable reference-level feedback from high-quality examples to guide the synthesis of new instructions and responses. DESIGNER \citep{DESIGNER} targets complex reasoning data synthesis by abstracting reusable design logic from existing high-quality questions and transferring it to new disciplinary materials or documents. Other recent methods, including GLAN \citep{li2025synthetic}, DS2-Instruct \citep{xu2026ds2}, and CrowdSelect \citep{li2026crowdselect}, respectively improve synthetic data generation from taxonomy-driven, domain-specific, and data-selection perspectives. These studies indicate that the effectiveness of synthetic data depends not only on scale, but also on transferable feedback signals.
\par
Beyond general instruction data synthesis, another line of work focuses on conflict scenarios in multi-turn interactions. MultiTurnInstruct \citep{multiturninstruct} constructs tasks involving multi-turn dependencies, information integration, and instruction conflicts to evaluate instruction following in complex multi-turn dialogue. Its contradiction resolution tasks further cover conflict-handling scenarios such as privacy protection, personalization, and prioritization. ConInstruct \citep{coninstruct} focuses on explicit conflicts within a single user instruction, constructing conflict examples across six constraint dimensions: content, keyword, phrase, length, format, and style. It evaluates models’ ability to detect and handle incompatible instruction requirements.
\par
Overall, existing data synthesis methods have advanced general dialogue generation, feedback-guided synthesis, and complex reasoning data construction. However, they provide limited support for extracting relevant information from dialogue history and constructing controllable examples of implicit conflicts between historical requirements and follow-up user utterances.

\section{Motivation}
\subsection{Definitions}
We \textbf{formulate user-side conflict detection as a three-way classification task}, where each current user utterance is classified based on its consistency with previously established user requirements: normal dialogue, explicit revision, and implicit conflict.
\par
\textbf{Normal dialogue} refers to cases where the current user utterance is consistent with the requirements that remain valid in the dialogue history. This category covers utterances that continue the original task, add compatible information under the original goal, or introduce a new task that does not affect existing requirements.
\par
\textbf{Explicit revision} refers to cases where the user explicitly updates, replaces, or negates a previously stated requirement within the same task. The revision may be expressed through an overt signal, such as ``change it to,'' ``not ... but ...,'' or ``ignore the previous requirement,'' or by directly specifying an incompatible replacement value.
\par
\textbf{Implicit conflict} refers to cases where the user introduces a new requirement within the same task that is inconsistent with requirements that remain valid in the dialogue history, without explicitly updating, replacing, or negating them. Such conflicts must instead be inferred from the task goal, dialogue context, or commonsense expectations.

\subsection{Preliminary Study}
\label{sec:preliminary}
To preliminarily assess the ability of LLMs to recognize user-side implicit conflicts, we manually construct WildChat-UC, a small-scale evaluation subset of UC-Bench introduced later.  Details of the construction process are provided in Appendix~\ref{appendix:uc-bench}. The resulting set covers both Chinese and English and contains three types of instances: implicit conflict, explicit revision, and normal dialogue. The statistics of WildChat-UC are summarized in Table~\ref{tab:uc-bench-statistics}.

We evaluate the selected LLMs under two protocols.

\textbf{Response-based evaluation.}
This protocol evaluates implicit conflict handling based on model responses. Given a dialogue history and a follow-up user utterance labeled as an implicit conflict, the evaluated LLM first generates a response without receiving any additional instructions. Subsequently, DeepSeek-V4-Flash is employed to evaluate whether the generated response accurately identifies the implicit conflict. A response is deemed correct only if it recognizes the presence of the conflict; otherwise, it is classified as incorrect. This evaluation protocol is exclusively applied to implicit conflict instances in WildChat-UC, with accuracy adopted as the evaluation metric.

\textbf{Classification-based evaluation.}
This protocol formulates the task as three-way classification. Given a dialogue history and the current user utterance, an LLM assigns the utterance to one of three categories: implicit conflict, explicit revision, or normal dialogue. We report Precision, Recall, and F1, together with Recall for the implicit conflict class.

\begin{table}[t]
\centering
\caption{\footnotesize Resp. Acc. denotes response-based accuracy. P, R, and F1 denote macro-precision, macro-recall, and macro-F1. R(IC) denotes recall on implicit conflict samples. All values are percentages.}
\label{tab:wildchat_uc_results}
\footnotesize
\setlength{\tabcolsep}{2.5pt}
\resizebox{\columnwidth}{!}{
\begin{tabular}{lccccc}
\toprule
\multirow{2}{*}{Model} 
& Resp. 
& \multicolumn{4}{c}{WildChat-UC} \\
\cmidrule(lr){2-2} \cmidrule(lr){3-6}
& Acc. 
& P 
& R 
& F1 
& R(IC) \\
\midrule
Claude Opus 4.8      & 28.33 & 70.91 & 70.00 & 61.71 & 40.00 \\
Claude Opus 4.7      & 21.67 & 67.69 & 66.67 & 57.59 & 35.00 \\
GPT-5.5              & 30.00 & 64.51 & 58.89 & 43.96 & 11.67 \\
GPT-5.4              & 25.00 & 70.54 & 58.33 & 45.46 & 10.00 \\
GLM-5.1              & 28.33 & 70.19 & 68.33 & 61.09 & 40.00 \\
Kimi K2.6            & 31.67 & 69.25 & 72.78 & 66.35 & 53.33 \\
DeepSeek-V4-Pro      & 11.67 & 62.21 & 57.78 & 45.21 & 18.33 \\
DeepSeek-V4-Flash    & 15.00 & 66.45 & 58.89 & 44.37 & 11.67 \\
Doubao-Seed-2.0-Pro  & 6.67  & 75.93 & 66.67 & 55.23 & 20.00 \\
Doubao-Seed-2.0-Lite & 8.33  & 66.55 & 62.22 & 46.97 & 11.67 \\
Doubao-Seed-2.0-Mini & 6.67  & 72.04 & 55.00 & 40.24 & 5.00  \\
Qwen3.7 Max          & 31.67 & 66.12 & 65.56 & 58.58 & 41.67 \\
Qwen3.7 Plus         & 30.00 & 68.77 & 73.33 & 65.80 & 50.00 \\
Qwen3.5-27B          & 18.33 & 68.39 & 64.44 & 56.57 & 28.33 \\
Qwen3.5-9B           & 11.67 & 67.78 & 63.33 & 48.79 & 15.00 \\
Qwen3.5-4B           & 11.67 & 54.40 & 52.78 & 34.77 & 8.33  \\
\bottomrule
\end{tabular}
}
\vspace{-0.5em}
\end{table}

We summarize the main findings as follows.

\textbf{Finding 1: LLMs rarely recognize implicit conflicts proactively during open-ended response generation.}
The response-based evaluation shows that, when LLMs are not explicitly instructed to detect implicit conflicts, they tend to directly follow the current user utterance rather than identify its potential incompatibility with prior requirements. As shown in Table~\ref{tab:wildchat_uc_results}, the 16 evaluated models achieve an average accuracy of only 19.79\% on implicit conflict instances, with the best model reaching 31.67\%. Moreover, 8 of the 16 models obtain an accuracy no higher than 20\%.

\textbf{Finding 2: LLMs show limited recall for implicit conflicts and tend to misclassify them as normal dialogue.}
Compared with the response-based evaluation, LLMs achieve better overall performance under the classification-based protocol, yet their ability to identify implicit conflicts remains limited. As shown in Table~\ref{tab:wildchat_uc_results}, the 16 evaluated models achieve an average IC recall of only 25.00\%, with the best model reaching 53.33\%; 9 of the 16 models achieve an IC recall no higher than 20\%. Although the average F1 reaches 52.04\%, this overall score masks the models' weak performance on the implicit conflict class. To further analyze the source of low recall, we present the confusion matrices of four representative models, Claude Opus 4.8, GPT-5.5, Kimi K2.6, and Qwen3.5-4B, in Figure~\ref{fig:confusion}. The results show that true implicit conflict instances are frequently misclassified as normal dialogue. This indicates that LLMs often fail to capture the implicit incompatibility between the current utterance and the dialogue history, instead treating it as a natural continuation consistent with the context.

\textbf{Finding 3: Response-based recognition is strongly associated with classification-based recognition, but the two are not equivalent.}
We further examine the relationship between response-based accuracy and IC recall under classification-based evaluation. As shown in Figure~\ref{fig:corr}, the two metrics are positively correlated across models, with Pearson's \(r = 0.7223\) (\(p = 0.0016\)) and Spearman's \(\rho = 0.6696\) (\(p = 0.0045\)). The corresponding linear regression yields \(R^2 = 0.5217\). These results indicate that models that more reliably recognize implicit conflicts during open-ended response generation also tend to identify them more accurately under classification. However, the two abilities should not be conflated, as their linear relationship explains only 52.17\% of the cross-model variation in response-based accuracy. Other factors, such as response strategies and instruction-following behavior, may also affect response-based recognition. Therefore, classification-based evaluation provides a complementary perspective on implicit conflict recognition and supports the development of a dedicated implicit conflict detector.

\textbf{Finding 4: LLMs struggle with implicit conflict recognition largely because they fail to identify the correct conflict rationale.}
During annotation, we retain a conflict rationale for each implicit conflict instance and use it as the gold rationale. To better understand the low IC recall, we examine whether the model-generated rationale covers the corresponding gold rationale.  Figure~\ref{fig:heatmap} shows the joint distribution of label predictions and rationale coverage for true implicit conflict instances. When the model-generated rationale misses the gold rationale, these instances are frequently misclassified as normal dialogue; when the gold rationale is covered, they are more likely to be correctly classified as implicit conflict. Figure~\ref{fig:coverage} further shows that incorrect predictions are concentrated among instances with missing rationales, indicating a strong association between rationale recovery and label correctness. Nevertheless, Qwen3.5-4B exhibits a different error pattern from the three closed-source models, as it sometimes covers the gold rationale but still predicts an incorrect label. This suggests that rationale coverage alone does not guarantee correct classification of the conflict relation.
\par
Overall, general-purpose LLMs show limited ability to proactively recognize implicit conflicts during response generation, reflecting their difficulty in identifying incompatibilities between the dialogue history and the current user utterance. Motivated by these findings, we propose SynUC, a constraint-guided data synthesis method for constructing targeted training data and training a lightweight LLM specialized in user-side conflict detection.

\section{Method}
\subsection{SynUC}
\label{sec:synuc}

\begin{figure*}
    \centering
    \includegraphics[width=1\linewidth]{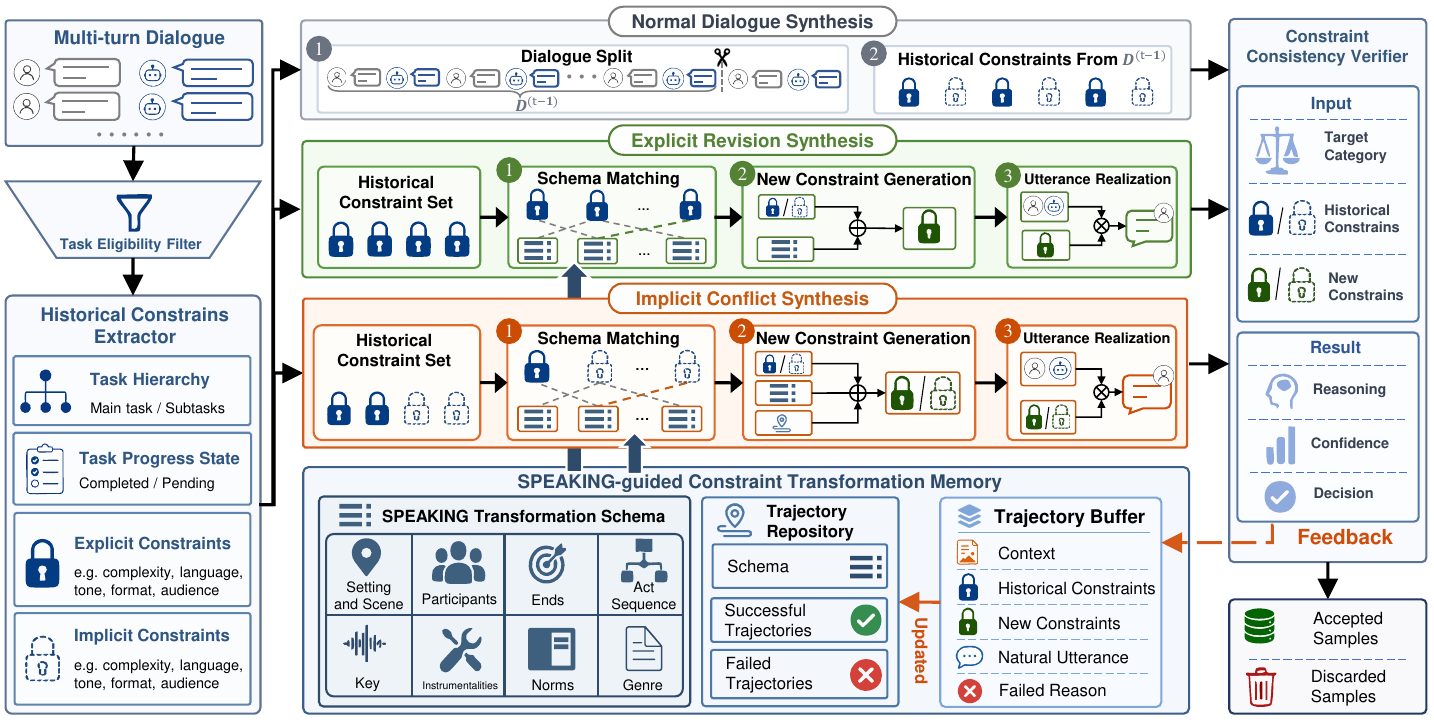}
    \caption{Overview of SynUC. Given a multi-turn dialogue, SynUC extracts the active historical constraints, constructs category-specific follow-up utterances through three synthesis pipelines, and verifies the synthesized samples using a constraint consistency verifier. Verified trajectories are further used to update the SPEAKING-guided constraint transformation memory.}
    \label{fig:method}
\end{figure*}

SynUC synthesizes label-controllable training samples for user-side conflict detection in a constraint space. As shown in Figure~\ref{fig:method}, it consists of four stages. First, a task eligibility filter determines whether a seed dialogue is suitable for constructing explicit revision and implicit conflict samples; the remaining dialogues are used for normal dialogue. Details of the task eligibility filter are provided in Appendix~\ref{appendix:task_eligibility_filter}. Second, a historical constraint extractor identifies the active task and its valid constraints. Third, SynUC constructs category-specific constraint relations and realizes them as natural follow-up user utterances. Finally, a two-stage verifier checks whether each synthesized sample satisfies its target category.
\par
\textbf{Because SynUC operates in a constraint space, we redefine the three categories based on the relations between historical constraints and the new constraints introduced by the follow-up user utterance.} \textbf{Explicit constraints} are requirements directly stated in the utterance, whereas \textbf{implicit constraints} are requirements inferred from the task context or commonsense knowledge. For example, ``English'' and ``formal'' are explicit constraints in ``Help me write a formal English job application email,'' whereas ``writing an academic paper'' may imply the implicit constraints of ``formality'' and ``rigor.''
\par
Under this representation, \textbf{explicit revision} denotes an incompatibility within the same task scope between a historical explicit constraint and a new explicit constraint. \textbf{Implicit conflict} denotes an incompatibility within the same task scope between historical and new constraints that involves at least one implicit constraint and contains no explicit revision signal. It covers historical implicit--new explicit, historical explicit--new implicit, and historical implicit--new implicit relations. \textbf{Normal dialogue} denotes cases where the new constraints remain compatible with all valid historical constraints within the same task scope.
\par
The complete SynUC procedure is provided in Algorithm~\ref{alg:synuc}, and its components are detailed below.

\subsubsection{Historical Constraint Extractor}
The historical constraint extractor maps a multi-turn dialogue into a structured representation of its task hierarchy and historical constraints, providing the basis for subsequent constraint revision and conflict construction. Given a multi-turn dialogue \(D^{(t)}=\{(u_i,a_i)\}_{i=1}^{t}\), where \(u_i\) and \(a_i\) denote the user utterance and the corresponding assistant response at turn \(i\), respectively, the extractor produces \(\mathcal{H}=\langle \mathcal{T}, \mathcal{C}\rangle\). Here, \(\mathcal{T}\) represents the task hierarchy, and \(\mathcal{C}\) contains the historical constraints that remain valid within the active task scope.
\par
SynUC first uses an LLM to identify task evolution in the dialogue and organize user goals into main tasks and subtasks. The resulting task hierarchy is represented as
\(
\mathcal{T}=
\langle
M_{\mathrm{done}},
M_{\mathrm{act}},
S_{\mathrm{done}},
S_{\mathrm{act}}
\rangle .
\)
Here, \(M_{\mathrm{done}}\) and \(S_{\mathrm{done}}\) denote completed main tasks and subtasks, while \(M_{\mathrm{act}}\) and \(S_{\mathrm{act}}\) denote the active main task and subtask. Main tasks capture relatively stable high-level user goals. When the user clearly initiates a new independent goal, the previous task is treated as completed. This hierarchy distinguishes task continuation from task switching, preventing constraints from completed tasks or subtasks from being incorrectly retained as active constraints.
\par
Conditioned on \(\mathcal{T}\), SynUC further uses an LLM to extract the valid constraints within the active task scope. Each constraint is represented as \(c=\langle \tau, d, v, \sigma, e \rangle\), where \(\tau\in\{\mathrm{explicit},\mathrm{implicit}\}\) is the constraint type, \(d\) is its dimension, \(v\) is its value, \(\sigma\in\{M_{\mathrm{act}},S_{\mathrm{act}}\}\) denotes its task scope, and \(e\) is the supporting dialogue evidence. The resulting historical constraint set is \(\mathcal{C}=\{c_j\}_{j=1}^{n}\).
\par
Finally, \(\mathcal{C}\) is partitioned into the explicit constraint set \(\mathcal{E}=\{c_j \in \mathcal{C} \mid \tau_j=\mathrm{explicit}\}\) and the implicit constraint set \(\mathcal{I}=\{c_j \in \mathcal{C} \mid \tau_j=\mathrm{implicit}\}\). Thus, \(\mathcal{C}=\mathcal{E}\cup\mathcal{I}\) and \(\mathcal{E}\cap\mathcal{I}=\varnothing\).

\subsubsection{SPEAKING-guided Constraint Transformation Memory}
To avoid relying on fixed manual templates for constraint conflict synthesis, SynUC introduces a SPEAKING-guided constraint transformation memory that accumulates transferable knowledge about constraint transformations. \textbf{The memory consists of three components: a SPEAKING transformation schema, a trajectory repository, and a trajectory buffer}. The schema guides both explicit revision and implicit conflict synthesis, whereas the repository and buffer are used only for implicit conflict synthesis, which is more challenging and benefits more from trajectory-level feedback.
\par
Inspired by Hymes's SPEAKING \citep{SPEAKING} model of communicative events, SynUC constructs a \textbf{SPEAKING transformation schema} to guide constraint changes across communicative dimensions. It treats a user utterance as a communicative event jointly governed by eight dimensions: setting and scene, participants, ends, act sequence, key, instrumentalities, norms, and genre, with detailed definitions provided in Appendix~\ref{appendix:speaking}. Rather than defining fixed transformation rules, the schema identifies the communicative dimensions along which constraints may change and guides the LLM to generate new constraints within the selected dimension. Formally, each schema characterizes a dimension-specific transformation relation between historical and new constraints. We define the schema set as $\mathcal{S}=\{s_k\}_{k=1}^{8}$, where $s_k$ denotes the $k$-th schema.
\par
For example, the \textit{participants} dimension can capture changes in the target audience, knowledge background, or user identity. Suppose the historical constraints require a text for general parents using plain language, while the new utterance asks for quantitative indicators and research hypotheses. Although the utterance does not explicitly revise the target audience, these requirements imply a shift toward expert reviewers and may therefore conflict with the historical audience constraint.
\par
SynUC further maintains a dynamic \textbf{trajectory repository} and \textbf{trajectory buffer} for each schema. Both use the same trajectory format, recording the context, historical constraints, generated new constraint, realized utterance, and verification outcome. For failed samples, it also records the failure reason. For each schema $s_k$, the trajectory repository is denoted as 
\begin{equation}
\mathcal{R}_{s_k}=\langle \mathcal{R}_{s_k}^{+}, \mathcal{R}_{s_k}^{-}\rangle.
\end{equation}
Here, $\mathcal{R}_{s_k}^{+}$ stores successful trajectories, while $\mathcal{R}_{s_k}^{-}$ stores failed trajectories together with their failure reasons. The repository retains representative successful and failed trajectories that provide reusable evidence and contrastive feedback for subsequent sample synthesis.
\par
The trajectory buffer temporarily stores newly generated trajectories before they are incorporated into the repository. For each schema $s_k$, the corresponding buffer is denoted as
\begin{equation}
\mathcal{B}_{s_k}=\langle \mathcal{B}_{s_k}^{+}, \mathcal{B}_{s_k}^{-}\rangle .
\end{equation}
After a candidate sample is processed by the constraint consistency verifier, its trajectory is written to the buffer of the corresponding schema according to the verification outcome, rather than being immediately merged into the repository. This delayed update reduces the influence of noisy individual feedback and enables batch-level comparison of trajectories under the same schema.
\par
When the buffer of a schema reaches a predefined update threshold, SynUC triggers a buffer-to-repository update. The memory first merges the new trajectories in the buffer with the existing trajectories in the repository to form a candidate set. It then uses an LLM to deduplicate similar trajectories and filter low-quality ones, retaining at most $K$ representative trajectories for each schema, with $K=3$ by default. This process is formalized as
$$
\mathcal{R}_{s_k}^{+}\leftarrow\mathrm{Update}\left(\mathcal{R}_{s_k}^{+}, \mathcal{B}_{s_k}^{+}\right),\quad
\mathcal{R}_{s_k}^{-}\leftarrow\mathrm{Update}\left(\mathcal{R}_{s_k}^{-}, \mathcal{B}_{s_k}^{-}\right),
$$
\vspace{-1.2em}
\begin{equation}
\left|\mathcal{R}_{s_k}^{+}\right|\le K,\quad
\left|\mathcal{R}_{s_k}^{-}\right|\le K.
\end{equation}

\subsubsection{Implicit Conflict Synthesis}
Given the historical constraint set $\mathcal{C}$, SynUC matches the extracted constraints to the SPEAKING transformation schemas. It uses LLM-based semantic matching to select the most appropriate historical constraint $c_h$ and its corresponding schema $s_k$. Based on the trajectory repository $\mathcal{R}_{s_k}$, SynUC generates a new constraint as
\begin{equation}
c_{t+1}
\sim
p_{\theta}
\left(
c
\mid
c_h,
s_k,
\mathcal{R}_{s_k},
\mathcal{C}
\right),
\end{equation}
where $p_{\theta}$ denotes the LLM-based constraint generator. The generated constraint is intended to be incompatible with $c_h$ within the same task scope, with at least one of the two constraints being implicit.

Conditioned on the historical dialogue context, SynUC then realizes $c_{t+1}$ as a natural follow-up utterance $u_{t+1}$. Rather than stating $c_{t+1}$ directly, the realization process implicitly conveys it through the introduced entities, requested actions, or situational conditions. Consequently, the utterance appears as a natural task continuation, content elaboration, or supplementary request without any explicit revision signal. The generated sample is subsequently passed to the constraint consistency verifier. Regardless of the verification result, its generation trajectory is stored in the trajectory buffer $\mathcal{B}_{s_k}$.

\subsubsection{Explicit Revision Synthesis}
For explicit revision synthesis, SynUC restricts the candidate anchors to the historical explicit constraint set $\mathcal{E}$. An LLM-based semantic matcher selects a historical anchor constraint $c_h$ together with a SPEAKING transformation schema $s_k$ that specifies an appropriate transformation dimension. Unlike implicit conflict synthesis, this process does not consult the trajectory repository. The candidate new constraint is generated by
\begin{equation}
c_{t+1}
\sim
p_{\theta}
\left(
c
\mid
c_h,
s_k,
\mathcal{E}
\right),
\end{equation}
where $p_{\theta}$ is the LLM-based constraint generator. The generated constraint is intended to be explicit and incompatible with $c_h$ within the same task scope.

SynUC then verbalizes $c_{t+1}$ as a natural follow-up utterance $u_{t+1}$ based on the dialogue history. The utterance conveys the revision either through an overt revision expression or by directly specifying a replacement value. The generated sample is subsequently passed to the constraint consistency verifier.

\subsubsection{Normal Dialogue Synthesis}
Given an original multi-turn dialogue \(D^{(t)}=\{(u_i,a_i)\}_{i=1}^{t}\), SynUC treats the final user utterance $u_t$ as the candidate follow-up and the preceding dialogue $D^{(t-1)}$ as its historical context. SynUC first applies the historical constraint extractor to $D^{(t-1)}$ to obtain the historical constraint representation $\mathcal{H}^{(t-1)}=\langle \mathcal{T}^{(t-1)}, \mathcal{C}^{(t-1)} \rangle$. It then feeds $u_t$ and the historical constraint set $\mathcal{C}^{(t-1)}$ into the constraint consistency verifier.

\subsubsection{Constraint Consistency Verifier}
Preliminary experiments indicate that verifying only the final synthesized dialogue is insufficient for reliably identifying implicit conflicts. We therefore provide the synthesis trajectory as additional evidence and introduce a constraint consistency verifier.
\par
The verifier consists of two stages. The first stage performs anchor-level consistency verification. For implicit conflict and explicit revision samples, the verifier first checks whether the historical anchor constraint \(c_h\) and the new constraint \(c_{t+1}\) satisfy the target category definition. It then checks whether \(u_{t+1}\) faithfully realizes this relation. This process is formalized as
\begin{equation}
V_{\mathrm{anchor}} = \mathbb{I}\left[\mathcal{J}_y(c_h, c_{t+1}, u_{t+1})=1\right],
\end{equation}
where \(\mathcal{J}_y(\cdot)\) is an LLM-instantiated predicate that determines whether the relation expressed by \((c_h, c_{t+1}, u_{t+1})\) satisfies the definition of category \(y\). Only samples with \(V_{\mathrm{anchor}}=1\) proceed to the next stage. Normal dialogue samples do not involve conflict or revision anchors and therefore directly enter the second stage.
\par
The second stage performs global-context consistency verification. The verifier uses an LLM to extract the current constraint set \(\hat{\mathcal{C}}\) from \(u_{t+1}\) and evaluates whether its relation to the historical constraint set \(\mathcal{C}\) satisfies the definition of category \(y\) under the full context:
\begin{equation}
V_{\text{global}} =
\mathbb{I}
\left[
G_y(\mathcal{C}, \hat{\mathcal{C}})=1
\right],
\end{equation}
where $G_y(\cdot)$ denotes a global constraint-relation judgment function over the full context and is instantiated by the LLM. This stage filters out samples that pass the anchor-level check but fail to preserve the target label under the full historical context.
\par
In addition to the binary decisions, the verifier outputs a confidence score \(q \in [0,100]\) for the category judgment based on the constraint-level evidence and full dialogue context. A candidate sample is accepted only if it passes all verification stages and its confidence score exceeds a predefined threshold \(\tau\); otherwise, it is discarded.

\subsection{Constraint-Attribution Reward}
Although SFT learns category decisions from annotated reasoning trajectories, label-level supervision may still produce correct predictions with misattributed rationales. A model may assign the correct label by exploiting superficial cues rather than identifying the constraint pair underlying the conflict or revision. Such shortcut reasoning reduces the interpretability of constraint-relation recognition. To address this issue, we apply GRPO-style \citep{guo2025deepseek} reinforcement learning after SFT and add a constraint-attribution term to the reward, encouraging the model to identify the key constraint pair that supports its category decision.
\par
For implicit conflict and explicit revision samples, the synthesis process records the historical anchor constraint and the new constraint. We define the attribution reward \(R_{\mathrm{attr}}\) to assess whether the model's reasoning correctly identifies both the historical anchor constraint and the new constraint, and accurately characterizes each constraint as explicit or implicit according to the synthesis annotations. Together with the label reward \(R_{\mathrm{label}}\) and format reward \(R_{\mathrm{fmt}}\), the final reward is defined as
\begin{equation} R = R_{\mathrm{label}} + \alpha R_{\mathrm{fmt}} + \mathbb{I}\left[ R_{\mathrm{fmt}}=1 \land R_{\mathrm{label}}=1 \right] \beta R_{\mathrm{attr}}, 
\end{equation}
where \(\alpha\) and \(\beta\) control the contributions of the format and attribution rewards, respectively. We set \(\alpha=0.2\) and \(\beta=0.4\) by default. The attribution reward is computed only for implicit conflict and explicit revision samples and contributes only when both the predicted label and output format are correct. For normal dialogue samples, we set \(R_{\mathrm{attr}}=0\).

\section{Experimental Setup}
\label{sec:experimental_setup}
\subsection{UC-Bench}
Following the procedure described in Appendix~\ref{appendix:uc-bench}, we construct WildChat-UC, LMSYS-UC, and ShareGPT-UC from WildChat \citep{zhao2024wildchat}, LMSYS \citep{zheng2024lmsys}, and ShareGPT \citep{sharegpt2023}, respectively. Together, these subsets form UC-Bench, a benchmark for detecting user-side conflicts. Unlike WildChat-UC, which is designed for a focused analysis of implicit conflict, LMSYS-UC and ShareGPT-UC use more balanced class distributions. Detailed statistics are reported in Table~\ref{tab:uc-bench-statistics}.

\begin{table}[t]
\caption{Statistics of UC-Bench. W-UC, L-UC, and S-UC denote WildChat-UC, LMSYS-UC, and ShareGPT-UC, respectively. ND, ER, and IC denote normal dialogue, explicit revision, and implicit conflict.}
\label{tab:uc-bench-statistics}
\centering
\footnotesize
\begin{tabular*}{\columnwidth}{@{\extracolsep{\fill}}lrrrr@{}}
\toprule
\textbf{Statistic} & \textbf{W-UC} & \textbf{L-UC} & \textbf{S-UC} & \textbf{Total} \\
\midrule
ND instances        & 20   & 34   & 33   & 87   \\
ER instances        & 20   & 33   & 33   & 86   \\
IC instances        & 60   & 33   & 34   & 127  \\
Chinese             & 51   & 0    & 52   & 103  \\
English             & 49   & 100  & 48   & 197  \\
Avg. messages       & 5.38 & 6.16 & 6.94 & 6.16 \\
Avg. turns          & 2.69 & 3.08 & 3.47 & 3.08 \\
Max. turns          & 9.00 & 9.00 & 10.00 & 10.00 \\
Avg. tokens         & 1512.05 & 815.57 & 1672.93 & 1333.52 \\
Max. tokens         & 8145 & 2453 & 5638 & 8145 \\
\bottomrule
\end{tabular*}
\end{table}

\subsection{UC-Data}
To support training for user-side conflict detection, we construct UC-Data from WildChat. We first sample 2,000 high-quality multi-turn dialogues that pass rule-based and LLM-based semantic filtering. We then apply SynUC to these dialogues, using DeepSeek-V4-Flash as the default LLM instantiation, and obtain 3,192 synthetic samples. Among them, 388 fail the verification stage, leaving 2,804 verified follow-up samples. After threshold filtering based on the confidence score $q$, with $q \geq 80$, we retain 2,487 high-quality training instances. To prevent data leakage, we conduct a deduplication check between UC-Data and UC-Bench and find no duplicated samples. Detailed statistics are reported in Table~\ref{tab:uc-data-statistics}.

\begin{table}[t]
\caption{Statistics of UC-Data. ND, ER, and IC denote normal dialogue, explicit revision, and implicit conflict, respectively. Turn statistics are computed over historical dialogue turns, and token counts exclude system prompts.}
\label{tab:uc-data-statistics}
\centering
\footnotesize
\setlength{\tabcolsep}{3.5pt}
\begin{tabular*}{\columnwidth}{@{\extracolsep{\fill}}lrrrr@{}}
\toprule
\textbf{Statistic} & \textbf{ND} & \textbf{ER} & \textbf{IC} & \textbf{Total} \\
\midrule
Samples         & 894    & 676    & 917    & 2487 \\
Chinese         & 424    & 202    & 309    & 935  \\
English         & 470    & 474    & 608    & 1552 \\
Avg. messages   & 6.02   & 7.20   & 7.23   & 6.79 \\
Avg. turns  & 3.01   & 3.60   & 3.61   & 3.39 \\
Max. turns  & 15     & 17     & 17     & 17   \\
Avg. tokens     & 1020.78 & 1733.42 & 1738.17 & 1479.00 \\
Max. tokens     & 7628   & 8165   & 8223   & 8223 \\
\bottomrule
\end{tabular*}
\end{table}
\par
To further assess label quality, we randomly sample 33, 33, and 34 instances from the normal dialogue, explicit revision, and implicit conflict categories, respectively, for human evaluation. Two reviewers independently assign one of the three labels to each sampled instance. Their annotations match the synthetic labels for 93\% and 95\% of the instances, respectively. The two reviewers achieve a raw agreement of 94\% and a Cohen's $\kappa$ of 0.909 on the three-way classification task.

\par
UC-Data preserves the intermediate constraints, labels, and verification records produced throughout the synthesis process. From it, we derive two training formats: UC-Data-SFT for supervised fine-tuning and UC-Data-GRPO for GRPO-style reinforcement training. In UC-Data-SFT, each instance is organized as user-assistant messages: the user message includes the task instruction, category definitions, dialogue history, and the follow-up utterance to be classified, while the assistant message provides structured reasoning over the task hierarchy, retained historical constraints, explicit and implicit constraints in the follow-up user utterance, constraint compatibility, and the final label. In UC-Data-GRPO, each instance uses the same input prompt as UC-Data-SFT, but replaces the assistant response with a reference annotation containing the gold label. For explicit revision and implicit conflict instances, the annotation further includes the historical and new constraints that trigger the revision or conflict, enabling the reward function to jointly evaluate label prediction and constraint attribution. More details are provided in the released GitHub repository.
\par
\subsection{Training Details}
We use Qwen3.5-4B as the default backbone and train it with the ms-swift \citep{zhao2025swift} framework. Training consists of two stages: supervised fine-tuning and reinforcement learning, both using LoRA \citep{hu2022lora} for parameter-efficient adaptation. In the SFT stage, we set the LoRA rank to 8, use a learning rate of \(1\times10^{-4}\), and train for 4 epochs. In the reinforcement learning stage, we use DR-GRPO \citep{liu2025understanding} as the default optimization algorithm, set the LoRA rank to 8, use a learning rate of \(5\times10^{-7}\), and train for 1 epoch with 4 sampled generations and a sampling temperature of 0.8. All training experiments are conducted on a single NVIDIA RTX A6000 GPU.
\par
\subsection{Evaluation Protocol}
We adopt a classification-based evaluation protocol to assess models' ability to detect user-side conflicts. Given a multi-turn dialogue history and the follow-up user utterance, the model must assign one of three labels: normal dialogue, explicit revision, or implicit conflict. We report macro-averaged Precision, Recall, and F1 to measure overall classification performance across the three categories. Since implicit conflict is the primary focus of this work, we additionally report recall for this category to measure how effectively models identify implicit user-side conflicts. The decoding temperature is set to 0.01 for stable evaluation.

\subsection{Baselines}
We consider two groups of baselines. The first group includes strong prompt-based LLMs, such as Claude Opus 4.8, GPT-5.5, DeepSeek-V4-Pro, GLM-5.1, and Qwen3.5-4B. These models are used to assess the ability of general-purpose LLMs to identify user-side conflicts under zero-shot and prompt-enhanced settings. The second group consists of representative data synthesis methods, including Baize\citep{xu2023baize}, MultiTurnInstruct\citep{multiturninstruct}, REFED\citep{REFED}, ConInstruct \citep{coninstruct}, and DESIGNER \citep{DESIGNER}. Since these methods were not originally designed for user-side conflict synthesis, we reproduce and adapt them to our task to obtain fair and comparable synthesis baselines.

\section{Experimental Results}
\subsection{Independent Evaluation}
Following the setup in Section~\ref{sec:experimental_setup}, we evaluate strong existing LLMs and models fine-tuned on data synthesized by different methods for user-side conflict detection. For our method, we run each experiment five times, report the average result, and provide 95\% confidence intervals. The results are shown in Table~\ref{tab:independent_eval}.

\begin{table*}[t]
\centering
\scriptsize
\setlength{\tabcolsep}{2.2pt}
\renewcommand{\arraystretch}{1.05}
\caption{Independent evaluation results on WildChat-UC, LMSYS-UC, ShareGPT-UC, and the overall test set. Prompting baselines are evaluated with direct inference. For data synthesis baselines, MTI and ConI denote MultiTurnInstruct and ConInstruct, respectively. SynUC results are averaged over five runs, and the half-widths of the 95\% confidence intervals are reported in the corresponding CI rows. $^\dagger$ denotes the use of the constraint-attribution reward $R_{\mathrm{attr}}$ during DR-GRPO training.}
\label{tab:independent_eval}
\resizebox{\textwidth}{!}{
\begin{tabular}{lcccccccccccccccc}
\toprule
\multirow{2}{*}{Method}
& \multicolumn{4}{c}{WildChat-UC}
& \multicolumn{4}{c}{LMSYS-UC}
& \multicolumn{4}{c}{ShareGPT-UC}
& \multicolumn{4}{c}{Overall} \\
\cmidrule(lr){2-5}
\cmidrule(lr){6-9}
\cmidrule(lr){10-13}
\cmidrule(lr){14-17}
& P & R & F1 & R(IC)
& P & R & F1 & R(IC)
& P & R & F1 & R(IC)
& P & R & F1 & R(IC) \\
\midrule
\multicolumn{17}{l}{\textit{Prompting baselines}} \\
Claude Opus 4.8 & 70.91 & 70.00 & 61.71 & 40.00 & 83.58 & 66.67 & 66.80 & 45.45 & 74.37 & 69.22 & 68.33 & 47.06 & 76.05 & 67.94 & 65.58 & 43.31 \\
Claude Opus 4.7 & 67.69 & 66.67 & 57.59 & 35.00 & 77.44 & 67.71 & 66.99 & 42.42 & 76.89 & 73.20 & 72.43 & 52.94 & 73.98 & 68.59 & 65.53 & 41.73 \\
GPT-5.5 & 64.51 & 58.89 & 43.96 & 11.67 & 75.45 & 57.58 & 50.68 & 9.09 & 75.74 & 63.52 & 55.79 & 11.76 & 71.74 & 60.27 & 50.47 & 11.02 \\
GPT-5.4 & 70.54 & 58.33 & 45.46 & 10.00 & 69.91 & 57.58 & 51.35 & 9.09 & 73.57 & 59.45 & 54.78 & 14.71 & 71.70 & 58.34 & 50.43 & 11.02 \\
GLM-5.1 & 70.19 & 68.33 & 61.09 & 40.00 & 83.64 & 73.74 & 73.53 & 48.48 & 81.16 & 75.28 & 73.79 & 47.06 & 77.92 & 72.46 & 69.34 & 44.09 \\
Kimi K2.6 & 69.25 & 72.78 & 66.35 & 53.33 & 76.49 & 68.75 & 68.16 & 45.45 & 85.44 & 81.13 & 81.20 & 67.65 & 77.41 & 74.22 & 72.29 & 55.12 \\
DeepSeek-V4-Pro & 62.21 & 57.78 & 45.21 & 18.33 & 75.27 & 63.64 & 60.45 & 24.24 & 77.59 & 62.48 & 56.58 & 14.71 & 71.85 & 61.74 & 54.63 & 18.90 \\
DeepSeek-V4-Flash & 66.45 & 58.89 & 44.37 & 11.67 & 76.60 & 57.61 & 53.12 & 15.15 & 77.25 & 61.38 & 58.17 & 23.53 & 73.50 & 58.75 & 51.61 & 15.75 \\
Doubao-Seed-2.0-Pro & 75.93 & 66.67 & 55.23 & 20.00 & 78.17 & 62.63 & 60.65 & 27.27 & 80.52 & 64.44 & 60.50 & 20.59 & 77.87 & 63.94 & 58.47 & 22.05 \\
Doubao-Seed-2.0-Lite & 66.55 & 62.22 & 46.97 & 11.67 & 78.74 & 63.70 & 60.32 & 21.21 & 80.28 & 68.45 & 64.36 & 23.53 & 75.17 & 64.33 & 56.81 & 17.32 \\
Doubao-Seed-2.0-Mini & 72.04 & 55.00 & 40.24 & 5.00 & 80.32 & 50.51 & 43.10 & 3.03 & 75.33 & 51.46 & 43.86 & 5.88 & 75.68 & 51.96 & 42.32 & 4.72 \\
Qwen3.7 Max & 66.12 & 65.56 & 58.58 & 41.67 & 77.46 & 63.67 & 62.34 & 33.33 & 77.79 & 70.29 & 68.63 & 41.18 & 74.07 & 67.02 & 63.90 & 39.37 \\
Qwen3.7 Plus & 68.77 & 73.33 & 65.80 & 50.00 & 77.54 & 61.65 & 60.12 & 30.30 & 80.39 & 74.30 & 72.47 & 44.12 & 75.18 & 70.27 & 67.14 & 43.31 \\
Qwen3.5-27B & 68.39 & 64.44 & 56.57 & 28.33 & 77.12 & 64.74 & 64.74 & 36.36 & 84.03 & 73.29 & 72.57 & 44.12 & 76.58 & 67.01 & 64.29 & 34.65 \\
Qwen3.5-9B & 67.78 & 63.33 & 48.79 & 15.00 & 79.15 & 60.61 & 54.02 & 9.09 & 67.93 & 56.39 & 51.96 & 17.65 & 71.46 & 59.78 & 51.83 & 14.17 \\
Qwen3.5-4B & 54.40 & 52.78 & 34.77 & 8.33 & 56.84 & 53.71 & 47.60 & 9.09 & 60.07 & 58.35 & 54.73 & 23.53 & 55.83 & 54.69 & 46.01 & 12.60 \\
\midrule
\multicolumn{17}{l}{\textit{Data synthesis baselines (Qwen3.5-4B backbone)}} \\
Baize (SFT) & 74.70 & 68.33 & 65.65 & 65.00 & 68.20 & 62.80 & 61.01 & 72.72 & 73.13 & 69.01 & 69.48 & 67.64 & 72.12 & 66.04 & 65.93 & 67.71 \\
Baize (DR-GRPO) & 75.00 & 68.89 & 66.46 & 66.67 & 64.86 & 60.78 & 59.09 & 66.67 & 76.20 & 73.02 & 73.21 & 70.59 & 72.39 & 67.20 & 66.88 & 67.72 \\
MTI (SFT) & 75.06 & 70.56 & 71.75 & 81.67 & 69.24 & 61.79 & 57.58 & 81.82 & 75.73 & 72.91 & 73.25 & 82.35 & 74.28 & 68.10 & 68.70 & 81.89 \\
MTI (DR-GRPO) & 75.67 & 72.22 & 73.15 & 81.67 & 68.15 & 60.78 & 55.88 & 81.82 & 77.53 & 74.90 & 75.13 & 85.29 & 74.76 & 68.74 & 69.26 & 82.68 \\
REFED (SFT) & 64.77 & 63.89 & 56.52 & 36.67 & 70.86 & 62.72 & 62.08 & 51.52 & 77.25 & 72.10 & 71.60 & 61.76 & 70.89 & 65.00 & 63.26 & 47.24 \\
REFED (DR-GRPO) & 66.36 & 66.11 & 59.56 & 43.33 & 69.09 & 62.72 & 62.02 & 48.48 & 76.65 & 73.08 & 72.75 & 64.71 & 70.92 & 66.44 & 64.87 & 50.39 \\
ConI (SFT) & 65.90 & 63.89 & 58.57 & 46.67 & 66.29 & 52.61 & 47.52 & 51.52 & 77.27 & 74.06 & 73.98 & 67.65 & 69.07 & 62.46 & 60.91 & 53.54 \\
ConI (DR-GRPO) & 70.12 & 66.11 & 60.62 & 48.33 & 64.07 & 50.59 & 46.30 & 42.42 & 74.22 & 71.93 & 71.31 & 79.41 & 67.92 & 61.44 & 59.76 & 55.12 \\
DESIGNER (SFT) & 67.75 & 65.00 & 59.96 & 50.00 & 68.35 & 60.67 & 58.46 & 54.55 & 81.79 & 78.02 & 77.91 & 76.47 & 73.11 & 67.12 & 65.85 & 58.27 \\
DESIGNER (DR-GRPO) & 67.87 & 64.44 & 60.88 & 58.33 & 70.54 & 60.70 & 59.20 & 54.55 & 81.63 & 79.06 & 78.79 & 73.53 & 73.87 & 68.18 & 67.16 & 61.42 \\
\midrule
\multicolumn{17}{l}{\textit{Our method (Qwen3.5-4B backbone)}} \\
SynUC (SFT) & 77.09 & 79.78 & 76.25 & 74.33 & 82.58 & 79.74 & 79.82 & 86.67 & 83.30 & 81.50 & 81.61 & 91.77 & 81.02 & 79.25 & 79.50 & 82.20 \\
\quad 95\% CI & $\pm$1.42 & $\pm$1.35 & $\pm$0.94 & $\pm$3.14 & $\pm$2.63 & $\pm$2.54 & $\pm$2.39 & $\pm$6.83 & $\pm$2.40 & $\pm$2.58 & $\pm$2.62 & $\pm$4.00 & $\pm$0.71 & $\pm$0.88 & $\pm$0.80 & $\pm$2.25 \\
SynUC (DR-GRPO) & 77.36 & 78.78 & 75.89 & 76.34 & 82.20 & 79.97 & 80.24 & 84.85 & 85.16 & 83.29 & 83.47 & 94.12 & 82.01 & 79.93 & 80.26 & 83.30 \\
\quad 95\% CI & $\pm$0.85 & $\pm$1.50 & $\pm$1.27 & $\pm$0.93 & $\pm$2.20 & $\pm$1.94 & $\pm$1.99 & $\pm$0.00 & $\pm$2.70 & $\pm$2.89 & $\pm$2.93 & $\pm$0.01 & $\pm$1.31 & $\pm$1.40 & $\pm$1.34 & $\pm$0.43 \\
SynUC (DR-GRPO$^\dagger$) & 79.77 & 81.67 & 78.53 & 77.00 & 83.56 & 80.95 & 81.27 & 84.85 & 84.68 & 82.47 & 82.57 & 95.88 & \textbf{82.69} & \textbf{80.66} & \textbf{81.01} & \textbf{84.09} \\
\quad 95\% CI & $\pm$1.22 & $\pm$0.85 & $\pm$0.59 & $\pm$3.40 & $\pm$1.81 & $\pm$2.31 & $\pm$2.12 & $\pm$3.76 & $\pm$2.10 & $\pm$2.28 & $\pm$2.33 & $\pm$3.27 & $\pm$0.65 & $\pm$0.73 & $\pm$0.59 & $\pm$1.07 \\
\bottomrule
\end{tabular}
}
\end{table*}

The overall results for the prompting baselines indicate that existing strong LLMs still struggle to reliably detect user-side implicit conflict. Some models achieve competitive overall classification performance: Kimi K2.6, GLM-5.1, and Qwen3.7 Plus obtain F1 scores of 72.29\%, 69.34\%, and 67.14\% on overall, respectively. However, their IC recall remains limited, at 55.12\%, 44.09\%, and 43.31\%. Other models perform substantially worse on this category: GPT-5.5, GPT-5.4, DeepSeek-V4-Pro, and Doubao-Seed-2.0-Mini achieve IC recall scores of only 11.02\%, 11.02\%, 18.90\%, and 4.72\%, respectively. These results indicate that direct prompting often overlooks incompatibilities involving implicit constraints, resulting in low IC recall.
\par
Models fine-tuned on SynUC-synthesized data consistently outperform prompting baselines, indicating that mapping user utterances from surface text into a constraint space and modeling the relation between historical and new constraints is effective for user-side conflict detection. In the SFT stage, SynUC achieves an overall F1 score of 79.50\% on UC-Bench, outperforming MultiTurnInstruct, the strongest SFT baseline, by 10.80 percentage points and ConInstruct by 18.59 percentage points. SynUC also shows stable performance across five runs, with the half-width of the 95\% confidence interval for overall F1 no larger than 0.8 percentage points. These results suggest that, compared with the predefined dimensions used by MultiTurnInstruct and ConInstruct, the SPEAKING communicative dimensions adopted by SynUC provide a more comprehensive and fine-grained characterization of user constraint changes, better aligning the synthesized data with human annotation criteria for user-side conflicts. In addition, REFED and DESIGNER both aim to abstract reusable experience from high-quality samples. However, their SFT F1 scores are 63.26\% and 65.85\%, respectively, both lower than that of Baize, which relies on self-chat data generation. This suggests that experience reuse alone does not guarantee better task adaptation. SynUC instead reuses verified high-quality constraint transformation trajectories, allowing experience to accumulate at the level of historical--new constraint relations and thereby generating training samples better aligned with user-side conflict detection.
\par
SynUC maintains a clear advantage under DR-GRPO, achieving an overall F1 score of 80.26\% on UC-Bench and outperforming MultiTurnInstruct (DR-GRPO), the strongest DR-GRPO baseline, by 11.00 percentage points. SynUC (DR-GRPO$^\dagger$), which further incorporates the constraint-attribution reward $R_{\mathrm{attr}}$, achieves 81.01\% F1 and 84.09\% IC recall on the overall test set, improving over the strongest baseline by 11.75 and 1.41 percentage points, respectively. These results indicate that constraint-level attribution signals provide additional benefits beyond label-level supervision for user-side conflict detection. SynUC also remains stable across five runs. For SynUC (DR-GRPO$^\dagger$), the half-widths of the 95\% confidence intervals for overall F1 and IC recall are 0.59 and 1.07 percentage points, respectively, suggesting that the gains are not due to single-run fluctuations.

\subsection{Pairwise Evaluation}
Beyond class labels, UC-Bench provides a human-annotated conflict rationale for each implicit conflict instance, specifying the historical and new constraints that give rise to the conflict. Classification metrics indicate whether a model predicts the correct label, but do not reveal whether its reasoning identifies the underlying constraint pair. We therefore conduct pairwise comparisons to evaluate constraint attribution quality. On the same implicit conflict instances, we compare SynUC (DR-GRPO$^{\dagger}$) with Baize, MultiTurnInstruct, REFED, ConInstruct, DESIGNER, and SynUC (DR-GRPO), using the human-annotated conflict rationale as the reference. Each comparison assesses which model more accurately identifies the conflicting constraints. Judgments are obtained from both an LLM judge and human annotators and are reported as Win, Tie, or Loss. Detailed evaluation protocols are provided in Appendix~\ref{sec:appendix-pairwise}.
\par
\begin{figure*}
    \centering
    \includegraphics[width=1\linewidth]{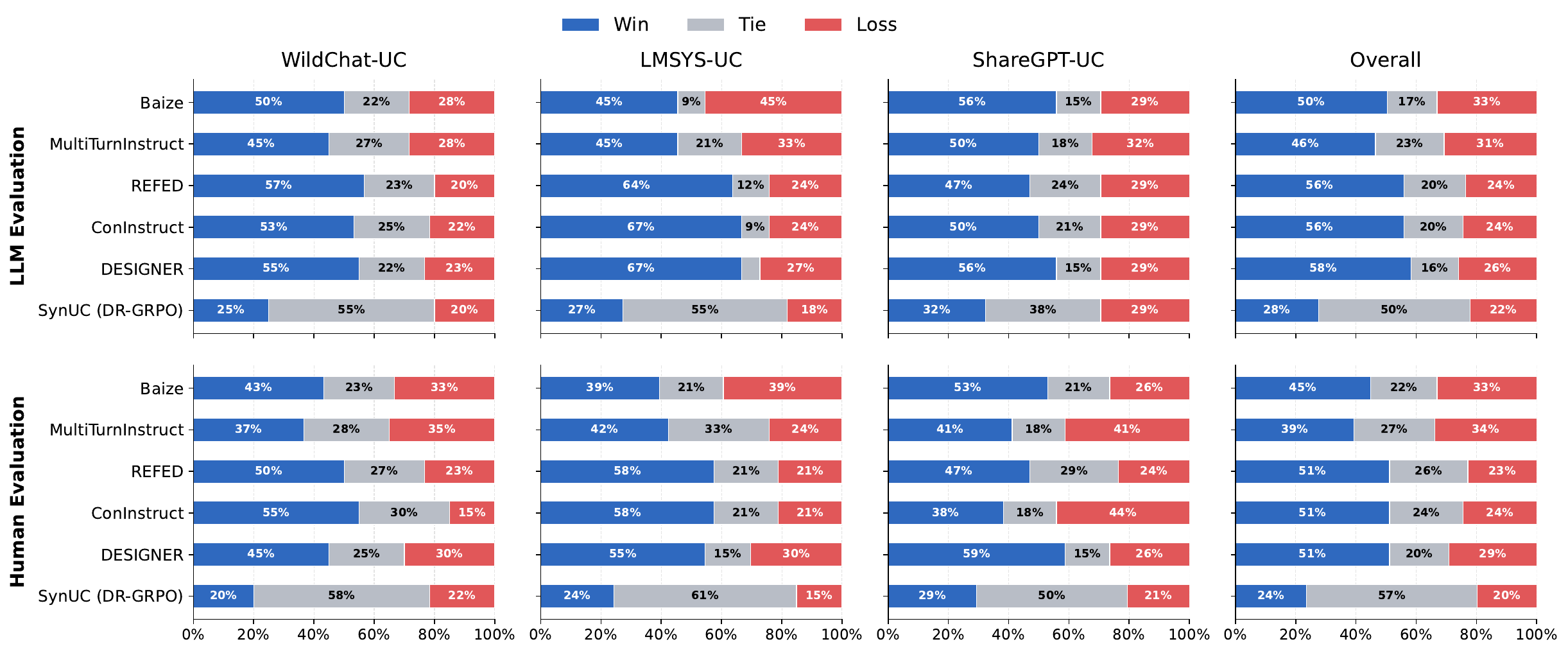}
    \caption{Pairwise comparison of constraint attribution quality on implicit conflict instances. Win, Tie, and Loss indicate whether SynUC (DR-GRPO$^\dagger$) outperforms, matches, or underperforms the compared baseline.}
    \label{fig:pairwise}
\end{figure*}
As shown in Figure~\ref{fig:pairwise}, under LLM-based evaluation, SynUC (DR-GRPO$^{\dagger}$) achieves a positive overall win--loss margin over every data synthesis baseline, with the most pronounced advantage on LMSYS-UC. This result suggests that SynUC-generated training data helps the model attribute implicit conflicts to the relevant constraint pairs, rather than relying solely on surface-level patterns for label prediction. Against SynUC (DR-GRPO), SynUC (DR-GRPO$^{\dagger}$) achieves overall Win, Tie, and Loss rates of 28\%, 50\%, and 22\%, respectively. The resulting positive win--loss margin indicates that $R_{\mathrm{attr}}$ further improves the model's ability to identify the key constraint pair underlying a conflict.
\par
Human evaluation yields the same overall trend. SynUC (DR-GRPO$^{\dagger}$) achieves a higher overall win rate than loss rate against every data synthesis baseline, although the win rates assigned by human annotators are generally lower than those assigned by the LLM judge. Against SynUC (DR-GRPO), its Win, Tie, and Loss rates are 24\%, 57\%, and 20\%, respectively. Together, these results provide further evidence that both SynUC-generated training data and the constraint-attribution reward $R_{\mathrm{attr}}$ improve constraint-level attribution for implicit conflicts.
\par
To assess evaluation reliability, we further report inter-annotator agreement and agreement between the human judgments and the LLM judgments. As shown in Figure~\ref{fig:annotator_agreement}, agreement between the two human annotators exceeds 80\% on the overall evaluation set, with Cohen's $\kappa$ above 0.70, indicating substantial inter-annotator agreement. As shown in Figure~\ref{fig:human_llm_agreement}, agreement between the human judgments and the LLM judgments exceeds 75\%, with Cohen's $\kappa$ above 0.60. These results indicate broad consistency between the LLM and human evaluations and support the use of the LLM judge for scalable evaluation.
\section{Discussion}
\subsection{Ablation Study}
To assess the contribution of each component in SynUC, we conduct ablation studies on its removable modules: task eligibility filtering, SPEAKING-guided constraint transformation memory, sample-level trajectory update, confidence filtering, and constraint consistency verification. We construct five ablation variants to examine whether these components improve the quality of synthesized samples. We remove trajectory update to evaluate the effect of sample-level experience updating. We remove memory and let the LLM generate conflict pairs directly, isolating the role of SPEAKING-guided constraint transformation knowledge in conflict construction. We remove task eligibility filtering to analyze the impact of unsuitable source dialogues on training data quality. We remove confidence filtering to measure the effect of low-confidence samples on model training. Finally, we replace constraint consistency verification with direct verification of the final sample, testing whether constraint-level verification is necessary.

\begin{figure*}[t]
    \centering
    \includegraphics[width=0.94\textwidth]{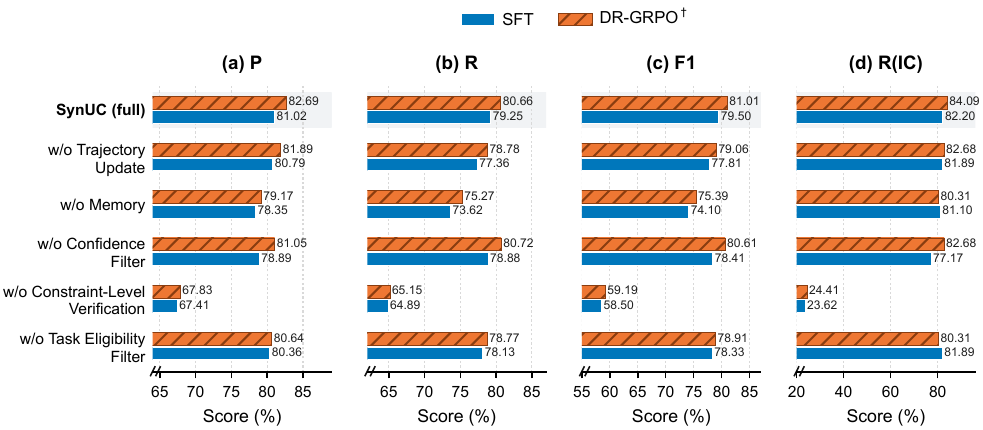}
    \caption{Ablation results of SynUC on UC-Bench under SFT and DR-GRPO$^\dagger$. Panels (a)--(d) report macro-precision, macro-recall, macro-F1, and recall for the implicit conflict class, respectively.}
    \label{fig:overall-ablation}
\end{figure*}

As shown in Figure~\ref{fig:overall-ablation}, removing constraint consistency verification leads to a substantial performance drop. Under DR-GRPO$^\dagger$, F1 decreases from 81.01\% to 59.19\%, and IC recall drops from 84.09\% to 24.41\%. The same trend appears under SFT. These results show that sample-level verification alone cannot reliably ensure label correctness.
\par
Removing the memory module also degrades performance. Under DR-GRPO$^\dagger$, the memory-ablated variant reduces F1 from 81.01\% to 75.39\%; under SFT, F1 drops from 79.50\% to 74.10\%. These results show that the SPEAKING-guided constraint transformation memory provides useful constraint transformation knowledge, rather than merely adding trajectory examples.
\par
Trajectory update provides stable but relatively moderate gains. Removing this module reduces  F1 under DR-GRPO$^\dagger$ from 81.01\% to 79.06\%, and IC recall from 84.09\% to 82.68\%. This finding suggests that sample-level trajectory updates may improve the robustness of conflict construction by accumulating verified successful and failed trajectories.
\par
Removing task eligibility filtering reduces F1 under DR-GRPO$^\dagger$ from 81.01\% to 78.91\%, and IC recall from 84.09\% to 80.31\%. This result indicates that not all source multi-turn dialogues are suitable for constructing user-side conflicts. By contrast, removing confidence filtering only slightly lowers F1 under DR-GRPO$^\dagger$ to 80.61\%, but reduces IC recall under SFT from 82.20\% to 77.17\%. This difference suggests that low-confidence samples have a more direct impact on supervised learning, whereas the subsequent DR-GRPO$^\dagger$ stage may partially mitigate their adverse effects.

\subsection{Effect of Assistant Response Retention}
Although this paper focuses on detecting conflicts among user-side constraints, a follow-up user utterance in multi-turn human--LLM dialogues often builds on the assistant's preceding response. Thus, even when the conflict arises between user constraints, assistant responses may affect the model's interpretation of contextual continuity. To examine their role in this task, we evaluate how different levels of assistant-response retention in the dialogue history influence model performance. Specifically, we compare three settings:
\begin{itemize}
\item \textbf{Full omission}: The content of each assistant response is removed, while a placeholder is retained to indicate an assistant turn.
\item \textbf{Partial omission}: The beginning and ending of each assistant response are retained, and the middle part is replaced with an omission marker. This is the default setting in this paper.
\item \textbf{Full retention}: Complete assistant responses are preserved, allowing the model to access the full interaction history.
\end{itemize}

\begin{table*}[t]
\centering
\scriptsize
\setlength{\tabcolsep}{3.0pt}
\renewcommand{\arraystretch}{1.08}
\caption{Effect of assistant-response retention on three UC-Bench subsets and the overall test set. P, R, F1, and R(IC) denote macro-precision, macro-recall, macro-F1, and recall on implicit conflict instances, respectively. All results are reported as percentages.}
\label{tab:assistant_retention}
\resizebox{\textwidth}{!}{
\begin{tabular}{>{\raggedright\arraybackslash}p{2cm} l cccc cccc cccc cccc}
\toprule
\multirow{2}{*}{Setting} & \multirow{2}{*}{Training}
& \multicolumn{4}{c}{WildChat-UC}
& \multicolumn{4}{c}{LMSYS-UC}
& \multicolumn{4}{c}{ShareGPT-UC}
& \multicolumn{4}{c}{Overall} \\
\cmidrule(lr){3-6}
\cmidrule(lr){7-10}
\cmidrule(lr){11-14}
\cmidrule(lr){15-18}
& & P & R & F1 & R(IC)
  & P & R & F1 & R(IC)
  & P & R & F1 & R(IC)
  & P & R & F1 & R(IC) \\
\midrule
\multirow{2}{*}{Full omission}
& SFT
& 76.92 & 79.07 & 75.94 & 78.89
& 81.00 & 78.64 & 78.86 & 84.85
& 79.27 & 76.61 & 76.70 & 82.35
& 79.98 & 77.57 & 77.68 & 81.37 \\
& DR-GRPO$^\dagger$
& 75.84 & 77.04 & 74.27 & 79.44
& 81.77 & 78.30 & 78.39 & 87.88
& 83.22 & 81.24 & 81.20 & 91.18
& 81.21 & 78.32 & 78.53 & 84.78 \\
\midrule
\multirow{2}{*}{Partial omission}
& SFT
& 77.09 & 79.78 & 76.25 & 74.33
& 82.58 & 79.74 & 79.82 & 86.67
& 83.30 & 81.50 & 81.61 & 91.77
& 81.02 & 79.25 & 79.50 & 82.20 \\
& DR-GRPO$^\dagger$
& 79.77 & 81.67 & 78.53 & 77.00
& 83.56 & 80.95 & 81.27 & 84.85
& 84.68 & 82.47 & 82.57 & 95.88
& 82.69 & 80.66 & 81.01 & 84.09 \\
\midrule
\multirow{2}{*}{Full retention}
& SFT
& 75.44 & 74.82 & 73.14 & 74.45
& 80.80 & 79.34 & 79.56 & 83.84
& 83.71 & 82.61 & 82.78 & 88.24
& 80.28 & 78.62 & 78.98 & 80.57 \\
& DR-GRPO$^\dagger$
& 75.98 & 76.85 & 74.48 & 75.55
& 79.81 & 78.34 & 78.59 & 82.83
& 82.57 & 80.56 & 80.77 & 91.18
& 79.93 & 77.94 & 78.37 & 81.63 \\
\bottomrule
\end{tabular}
}
\end{table*}
Table~\ref{tab:assistant_retention} reports the results under different assistant-response retention strategies. Partial omission achieves the best or near-best performance in most settings and provides the most stable performance on the overall test set. Compared with partial omission, full omission reduces overall F1 by 1.82 percentage points after SFT and by 2.48 percentage points after reinforcement learning. These results suggest that assistant responses help maintain contextual continuity in conflict detection. However, full retention also decreases overall F1 by 0.52 and 2.64 percentage points under the two training settings, respectively, indicating that retaining complete assistant responses provides no additional benefit and may introduce contextual redundancy. Partial omission therefore achieves a better balance between preserving relevant context and reducing context overhead.

\subsection{Schema Retrieval Strategy Analysis}
In SynUC, synthesizing implicit conflict and explicit revision samples requires retrieving an appropriate schema from memory based on historical constraints. We use LLM-guided retrieval by default, where the LLM selects the schema conditioned on the historical constraints and schema descriptions. To evaluate this design, we compare it against two alternatives. Lexical retrieval represents historical constraints and schema descriptions as text and selects the most relevant schema using BM25 lexical similarity. Dense retrieval encodes them with Qwen3-Embedding-0.6B and selects the schema with the highest cosine similarity. The results are reported in Table~\ref{tab:schema_retrieval}.

\begin{table*}[t]
\centering
\scriptsize
\setlength{\tabcolsep}{3.0pt}
\renewcommand{\arraystretch}{1.08}
\caption{Effect of schema retrieval strategies on three UC-Bench subsets and the overall test set. P, R, F1, and R(IC) denote macro-precision, macro-recall, macro-F1, and recall on implicit conflict instances, respectively. All results are reported as percentages.}
\label{tab:schema_retrieval}
\resizebox{\textwidth}{!}{
\begin{tabular}{>{\raggedright\arraybackslash}p{1.6cm} l cccc cccc cccc cccc}
\toprule
\multirow{2}{*}{Retrieval} & \multirow{2}{*}{Training}
& \multicolumn{4}{c}{WildChat-UC}
& \multicolumn{4}{c}{LMSYS-UC}
& \multicolumn{4}{c}{ShareGPT-UC}
& \multicolumn{4}{c}{Overall} \\
\cmidrule(lr){3-6}
\cmidrule(lr){7-10}
\cmidrule(lr){11-14}
\cmidrule(lr){15-18}
& & P & R & F1 & R(IC)
  & P & R & F1 & R(IC)
  & P & R & F1 & R(IC)
  & P & R & F1 & R(IC) \\
\midrule
\multirow{2}{*}{LLM-guided}
& SFT
& 77.09 & 79.78 & 76.25 & 74.33
& 82.58 & 79.74 & 79.82 & 86.67
& 83.30 & 81.50 & 81.61 & 91.77
& 81.02 & 79.25 & 79.50 & 82.20 \\
& DR-GRPO$^\dagger$
& 79.77 & 81.67 & 78.53 & 77.00
& 83.56 & 80.95 & 81.27 & 84.85
& 84.68 & 82.47 & 82.57 & 95.88
& 82.69 & 80.66 & 81.01 & 84.09 \\
\midrule
\multirow{2}{*}{Lexical}
& SFT
& 80.18 & 76.85 & 73.28 & 68.89
& 83.60 & 76.50 & 77.11 & 74.75
& 80.55 & 76.65 & 76.50 & 78.43
& 81.66 & 76.03 & 75.69 & 72.97 \\
& DR-GRPO$^\dagger$
& 81.81 & 80.37 & 77.12 & 74.44
& 82.84 & 77.19 & 77.39 & 79.80
& 83.13 & 78.61 & 78.32 & 84.31
& 83.13 & 77.87 & 77.68 & 78.48 \\
\midrule
\multirow{2}{*}{Dense}
& SFT
& 81.22 & 79.26 & 76.66 & 76.11
& 80.98 & 77.28 & 76.99 & 92.93
& 80.77 & 75.61 & 75.75 & 81.37
& 81.33 & 76.18 & 76.32 & 81.89 \\
& DR-GRPO$^\dagger$
& 79.95 & 80.37 & 77.24 & 72.78
& 79.72 & 75.60 & 75.26 & 90.91
& 80.48 & 76.25 & 76.30 & 85.29
& 80.32 & 75.71 & 75.91 & 80.84 \\
\bottomrule
\end{tabular}
}
\end{table*}
The results show that LLM-guided retrieval is the most stable strategy. After SFT, it outperforms lexical and dense retrieval in overall F1 by 3.81 and 3.18 percentage points, respectively. After DR-GRPO$^\dagger$, the corresponding differences are 3.33 and 5.10 percentage points. LLM-guided retrieval also achieves the highest overall IC recall, exceeding lexical and dense retrieval by 9.23 and 0.31 percentage points after SFT, and by 5.61 and 3.25 percentage points after DR-GRPO$^\dagger$, respectively. In contrast, lexical retrieval is constrained by surface-level matching and exhibits limited recall. Dense retrieval achieves high IC recall on some subsets, such as LMSYS-UC, but yields lower F1, suggesting that relying solely on vector similarity may lead to imbalanced class discrimination.

\subsection{Backbone Generalization Analysis}
\begin{figure}
    \centering
    \includegraphics[width=1\linewidth]{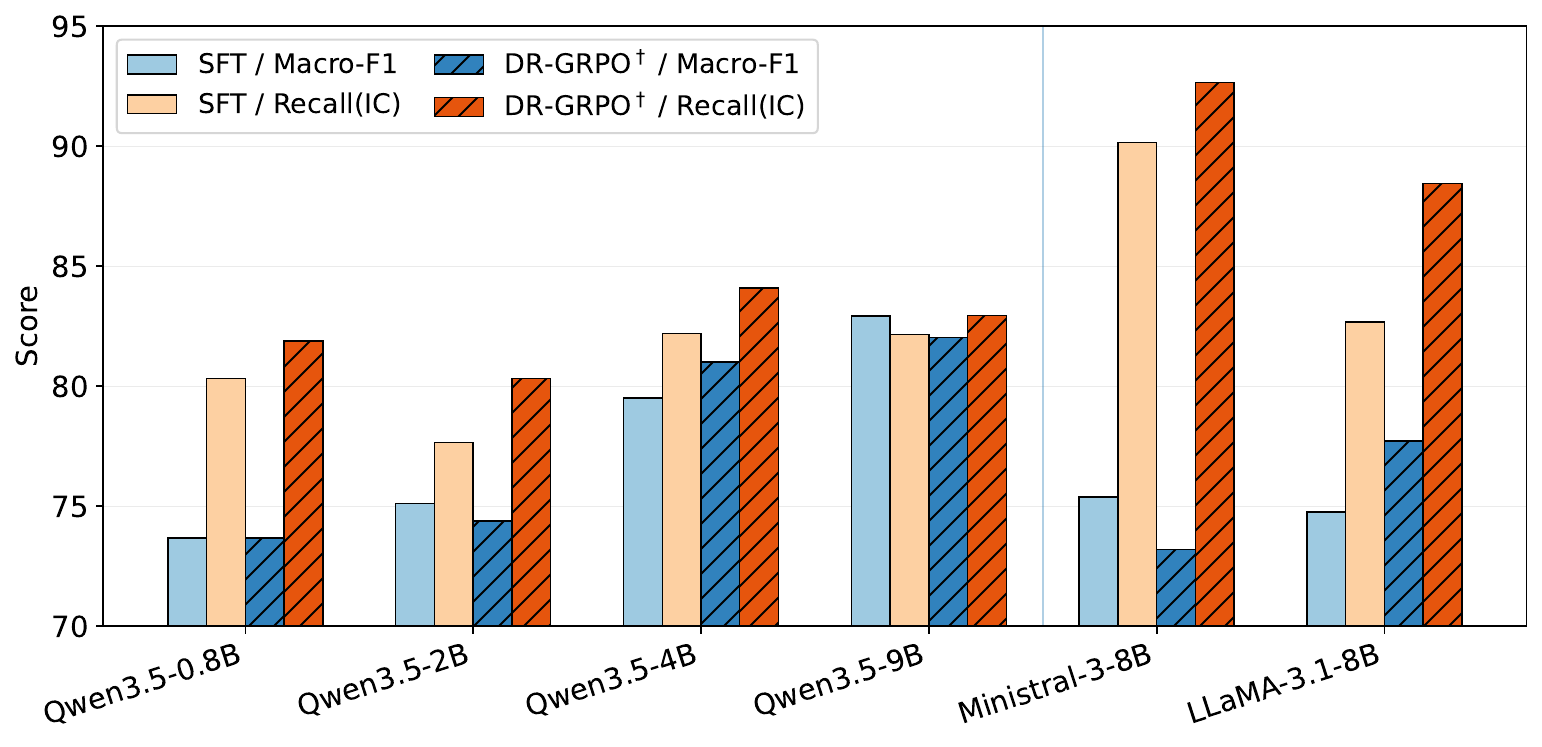}
    \caption{Backbone comparison on UC-Bench. Bars report macro-F1 and IC recall under SFT and DR-GRPO$^\dagger$.}
    \label{fig:backbone}
\end{figure}

To examine whether UC-Data can consistently improve user-side conflict detection across different model architectures and scales, we further evaluate multiple backbone models trained on the same dataset. Specifically, we select Qwen3.5 models with 0.8B, 2B, 4B, and 9B parameters, as well as Ministral-3-8B and LLaMA-3.1-8B, as backbone models. The results are summarized in Figure~\ref{fig:backbone}.

Overall, UC-Data yields stable improvements in user-side conflict detection across model families and parameter scales, and consistently outperforms the unfine-tuned large models reported in Table~\ref{tab:independent_eval}. This indicates that the dataset is not overly tied to a specific backbone. Within the Qwen3.5 series, larger models generally achieve stronger overall performance, with the 4B and 9B variants substantially outperforming the 0.8B and 2B variants in F1. This suggests that user-side conflict detection continues to benefit from stronger semantic understanding. Meanwhile, although LLaMA-3.1-8B and Ministral-3-8B lag behind Qwen3.5-4B and Qwen3.5-9B in F1, they still achieve high IC recall, with several results exceeding 90\%. These findings suggest that, even under the same UC-Data training setup, the initial capability differences among backbone models continue to shape their performance on user-side conflict detection.

\subsection{Effect of LLM Choice in SynUC}
This section examines how the choice of LLM affects data synthesis in SynUC. Beyond the default DeepSeek-V4-Flash, we replace the generator LLM with Doubao-Seed-2.0-Lite, MiMo-V2.5, and Qwen3.7 Plus, and evaluate the models trained on the resulting synthetic data.

\begin{table*}[t]
\centering
\scriptsize
\setlength{\tabcolsep}{3.0pt}
\renewcommand{\arraystretch}{1.08}
\caption{Effect of different generator LLMs in SynUC on UC-Bench. P, R, F1, and R(IC) denote macro-precision, macro-recall, macro-F1, and recall on implicit conflict instances, respectively. All results are reported as percentages.}
\label{tab:synthesis_llm}
\resizebox{\textwidth}{!}{
\begin{tabular}{>{\raggedright\arraybackslash}p{2.2cm} l cccc cccc cccc cccc}
\toprule
\multirow{2}{*}{Generator LLM} & \multirow{2}{*}{Training}
& \multicolumn{4}{c}{WildChat-UC}
& \multicolumn{4}{c}{LMSYS-UC}
& \multicolumn{4}{c}{ShareGPT-UC}
& \multicolumn{4}{c}{Overall} \\
\cmidrule(lr){3-6}
\cmidrule(lr){7-10}
\cmidrule(lr){11-14}
\cmidrule(lr){15-18}
& & P & R & F1 & R(IC)
  & P & R & F1 & R(IC)
  & P & R & F1 & R(IC)
  & P & R & F1 & R(IC) \\
\midrule
\multirow{2}{*}{DeepSeek-V4-Flash}
& SFT
& 77.09 & 79.78 & 76.25 & 74.33
& 82.58 & 79.74 & 79.82 & 86.67
& 83.30 & 81.50 & 81.61 & 91.77
& 81.02 & 79.25 & 79.50 & 82.20 \\
& DR-GRPO$^\dagger$
& 79.77 & 81.67 & 78.53 & 77.00
& 83.56 & 80.95 & 81.27 & 84.85
& 84.68 & 82.47 & 82.57 & 95.88
& 82.69 & 80.66 & 81.01 & 84.09 \\
\midrule
\multirow{2}{*}{Doubao-Seed-2.0-Lite}
& SFT
& 72.01 & 69.81 & 65.83 & 62.78
& 82.48 & 70.40 & 70.99 & 55.56
& 83.54 & 78.65 & 78.95 & 80.39
& 79.76 & 73.06 & 72.53 & 65.62 \\
& DR-GRPO$^\dagger$
& 71.32 & 70.93 & 66.65 & 64.44
& 81.74 & 69.73 & 70.21 & 55.56
& 82.78 & 77.64 & 78.01 & 80.39
& 79.02 & 72.81 & 72.24 & 66.40 \\
\midrule
\multirow{2}{*}{MiMo-V2.5}
& SFT
& 72.18 & 71.67 & 68.00 & 56.67
& 76.24 & 68.08 & 68.06 & 49.50
& 77.60 & 71.13 & 71.24 & 57.84
& 75.94 & 70.35 & 69.49 & 55.12 \\
& DR-GRPO$^\dagger$
& 72.11 & 72.04 & 68.01 & 56.11
& 77.06 & 68.09 & 68.28 & 50.50
& 77.92 & 73.77 & 73.61 & 62.74
& 76.20 & 71.17 & 70.30 & 56.43 \\
\midrule
\multirow{2}{*}{Qwen3.7 Plus}
& SFT
& 68.94 & 75.19 & 69.27 & 60.56
& 73.75 & 69.85 & 69.91 & 51.51
& 78.96 & 78.08 & 78.02 & 69.61
& 74.50 & 74.28 & 73.02 & 60.63 \\
& DR-GRPO$^\dagger$
& 70.35 & 75.93 & 70.77 & 62.78
& 73.05 & 69.51 & 69.42 & 50.50
& 78.52 & 77.41 & 77.39 & 69.61
& 74.65 & 74.29 & 73.15 & 61.42 \\
\bottomrule
\end{tabular}
}
\end{table*}

Table~\ref{tab:synthesis_llm} shows that the choice of LLM substantially affects downstream training performance. DeepSeek-V4-Flash achieves the best overall results. Models trained on data synthesized by Doubao-Seed-2.0-Lite and Qwen3.7 Plus also achieve stable performance and outperform direct inference with the corresponding LLMs, as shown in Table~\ref{tab:independent_eval}. These results indicate that the gains are not solely attributable to the inference capability of the synthesis LLM, but largely arise from SynUC's modeling of revision and conflict relations in the constraint space. In contrast, MiMo-V2.5 produces weaker overall performance, with a particularly noticeable drop in IC recall. This suggests that insufficient modeling of implicit incompatibilities between historical and new constraints can reduce the coverage of implicit conflict patterns in the synthetic data. Overall, SynUC is effective across different synthesis LLMs, and models trained on its synthesized data outperform direct inference with the corresponding LLMs in user-side conflict detection.

\subsection{Decision State Visualization}
To examine how different training stages affect the model's internal classification process, we visualize its decision states. For each sample in UC-Bench, we first prompt the LLM to generate a reasoning process. We then extract the hidden state immediately before the final category label is generated. This position captures the model's internal representation after task-level analysis, historical constraint identification, new constraint identification, and constraint compatibility assessment, and is therefore treated as a reasoning-conditioned decision state. We use UMAP \citep{McInnes2018} to project these states into two dimensions and color each sample according to its gold label. We also compute the silhouette score in the original high-dimensional space to quantify the separation among the three classes.

\begin{figure*}
    \centering
    \includegraphics[width=1\linewidth]{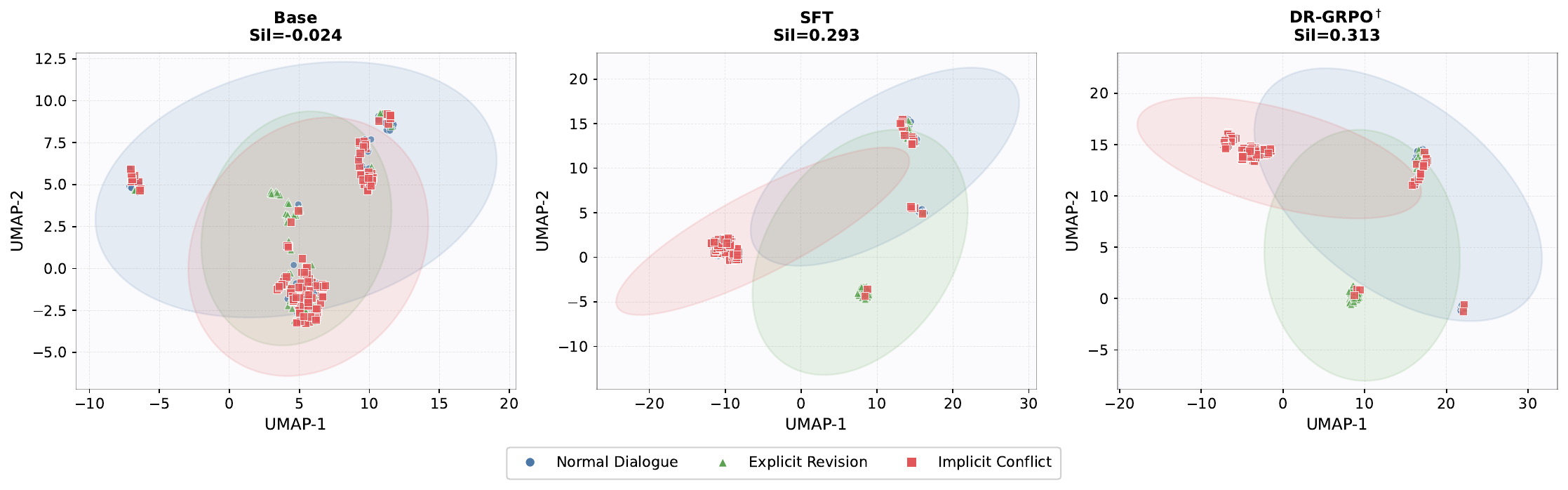}
    \caption{UMAP visualization of decision states across training stages.}
    \label{fig:visualization}
\end{figure*}

As shown in Figure~\ref{fig:visualization}, the Base model produces a mixed distribution with no clear class structure. Its silhouette score is -0.024, indicating that the decision states are not well aligned with the user-side constraint relation labels. After SFT, samples exhibit more pronounced class-dependent clustering, and the silhouette score increases to 0.293, suggesting that supervised fine-tuning helps form more discriminative decision states. DR-GRPO$^\dagger$ further improves the score to 0.313, indicating that reinforcement learning strengthens the alignment between decision states and target categories. Although some local overlap remains, the overall results show that SynUC training yields clearer representations of user-side constraint relations, thereby improving discrimination among the three classes.

\subsection{Case Study}
To illustrate the behavioral differences of Qwen3.5-4B before and after fine-tuning on UC-Data, we conduct a case study on a representative travel-planning example, as shown in Appendix Table~\ref{tab:case_study}.
\par
Before fine-tuning, Qwen3.5-4B notices that the newly requested activities may be less suitable for children. However, the surface-level additive cue ``also add'' leads it to interpret the request as a natural extension of the existing itinerary. It therefore fails to examine whether the new requirements conflict with still-valid historical constraints. This phenomenon is also supported by Finding~4 in Section~\ref{sec:preliminary}. Consequently, it classifies the instance as normal dialogue.
\par
After fine-tuning on UC-Data, Qwen3.5-4B correctly recovers the constraint-level conflict. It identifies the explicit historical constraints of ``avoiding crowded tourist attractions'' and selecting ``quiet places where kids can enjoy'', and further infers the implicit requirement of ``family-friendly activities''. It then extracts the newly introduced requirements of a ``wine-pairing dinner and late-night jazz show'', and infers their implicit property as ``adult-oriented or couple-oriented activities''. Based on this analysis, the model recognizes an incompatibility between the historical implicit constraint and the newly introduced implicit constraint, and correctly predicts the instance as implicit conflict.

\section{Conclusion}
This paper presents the first study of user-side implicit conflict detection in human--LLM dialogues and introduces UC-Bench, a human-annotated benchmark for evaluating this task. We further propose SynUC, a constraint-guided data synthesis method, and use it to construct UC-Data for model training. Experiments on UC-Bench show that models trained on UC-Data substantially outperform existing synthesis baselines and strong commercial LLMs, particularly in recognizing implicit conflicts. Further analyses validate the effectiveness of the components of SynUC. We also examine key design choices, including the synthesis LLM, backbone model, and retrieval strategy, and identify effective configurations for SynUC. Finally, we release UC-Bench, UC-Data, and the implementation of SynUC to support future research.
\par
This work has several limitations. First, UC-Bench can be further expanded to cover more task scenarios, languages, and long-dialogue settings. Second, future work can refine the taxonomy of implicit conflicts by constructing instances with different difficulty levels, enabling a more comprehensive evaluation of model capabilities. Finally, SynUC still relies on LLMs for constraint extraction, transformation, and verification. Future work can incorporate human-in-the-loop quality control with human feedback and iterative validation to improve the reliability of synthetic data.

\printcredits

\bibliographystyle{model5-names}

\bibliography{cas-refs}

\clearpage
\twocolumn[
\begin{center}
    {\Large\bfseries Appendix}
\end{center}
\vspace{1em}
]

\appendix
\section{Additional Results from the Preliminary Study}
\begin{figure*}
    \centering
    \includegraphics[width=1\linewidth]{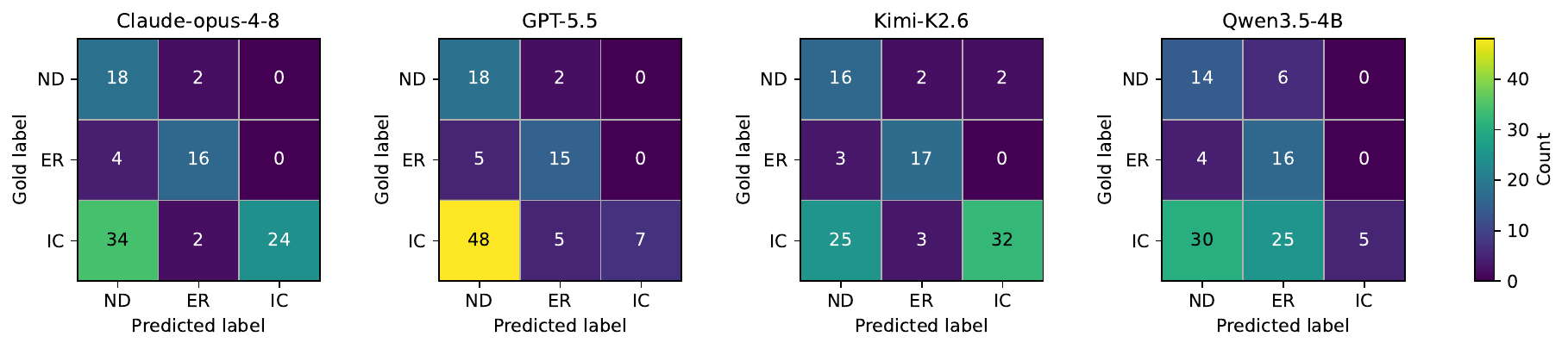}
    \caption{Confusion matrices of four representative LLMs on WildChat-UC under the classification-based evaluation. Rows denote gold labels and columns denote predicted labels.}
    \label{fig:confusion}
\end{figure*}

\paragraph{Confusion matrices.}
Figure~\ref{fig:confusion} presents the confusion matrices of four representative LLMs on WildChat-UC. The matrices offer a detailed view of the error patterns behind the classification results in Table~\ref{tab:wildchat_uc_results}. 

\begin{figure*}
    \centering
    \includegraphics[width=1\linewidth]{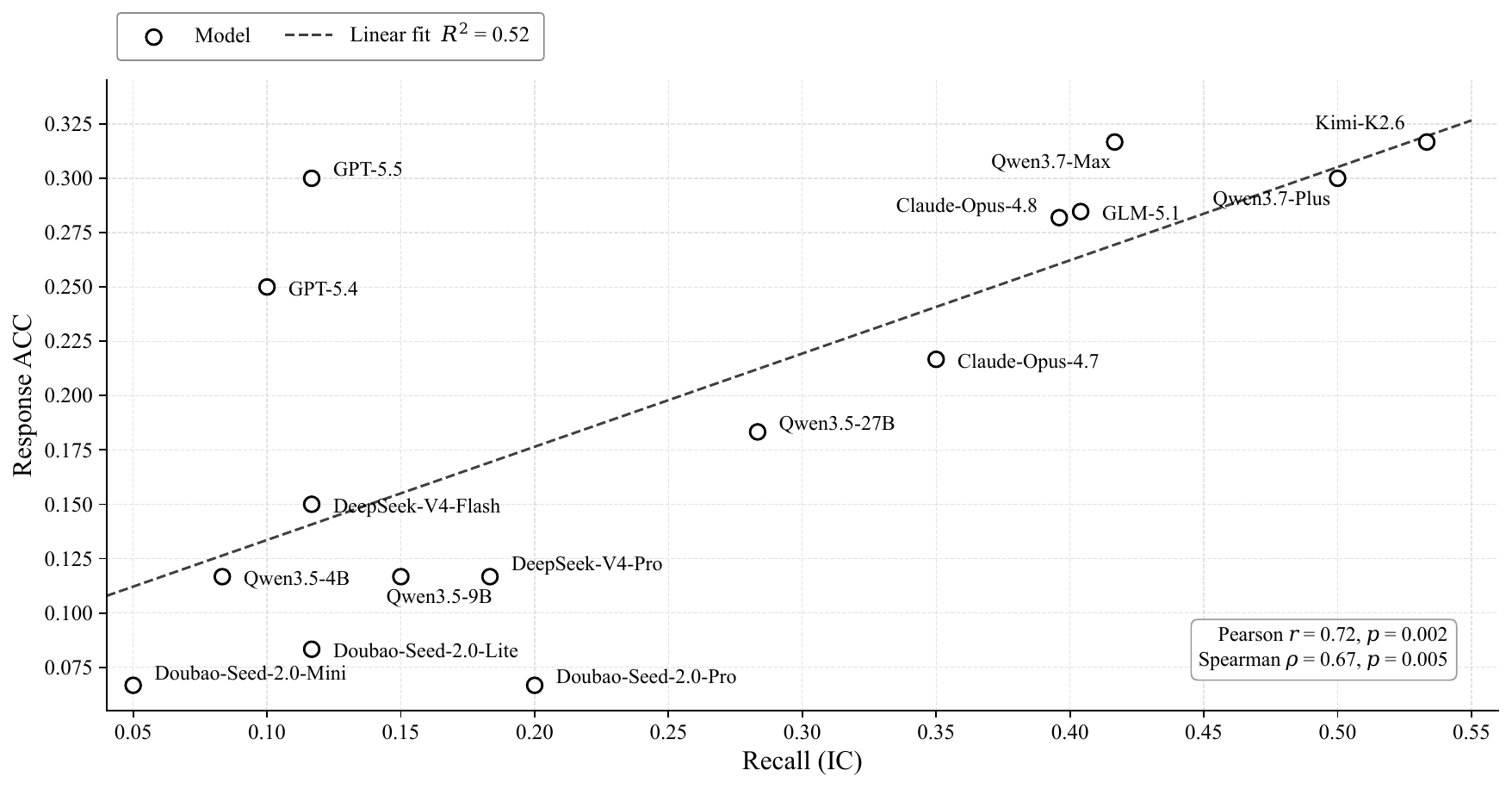}
    \caption{Correlation between classification-based implicit conflict recall and response-based accuracy. Each point denotes one evaluated model. The x-axis reports Recall (IC) in the classification setting, and the y-axis reports response-based accuracy on implicit conflict instances. The dashed line shows a linear fit.}
    \label{fig:corr}
\end{figure*}

\paragraph{Correlation analysis.}
Figure~\ref{fig:corr} shows the relationship between IC recall in the classification-based evaluation and response-based accuracy. 

\begin{figure*}
    \centering
    \includegraphics[width=1\linewidth]{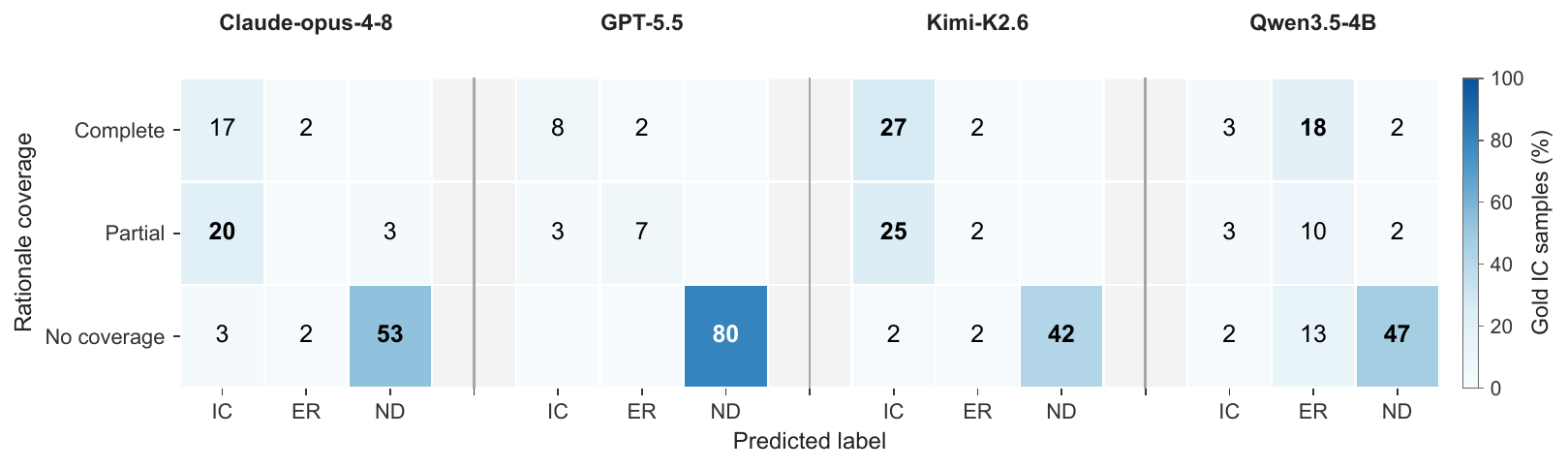}
    \caption{Joint distribution of predicted labels and rationale coverage on gold implicit conflict instances. Each panel corresponds to one representative model. Rows indicate the degree to which the model reasoning covers the gold conflict rationale, and columns indicate the predicted label. Cell values denote the percentage of gold implicit conflict samples.}
    \label{fig:heatmap}
\end{figure*}

\begin{figure*}
    \centering
    \includegraphics[width=0.8\linewidth]{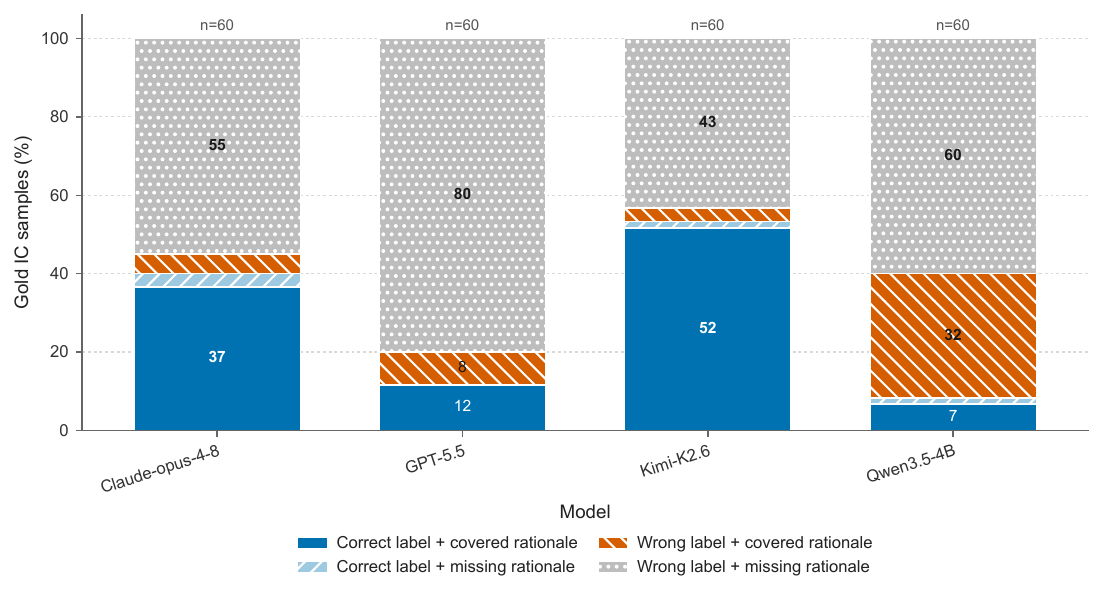}
    \caption{Distribution of label correctness and rationale coverage on gold implicit conflict instances. Each stacked bar corresponds to one representative model. Rationale coverage is computed by merging complete and partial coverage, and missing rationale corresponds to no coverage. Segment values denote percentages, with $n=60$ for each model.}
    \label{fig:coverage}
\end{figure*}

\paragraph{Rationale coverage analysis.}
We provide additional analyses of whether model reasoning covers the annotated conflict rationale on gold implicit conflict instances. For each instance, the gold rationale records the key reason that makes the current utterance incompatible with the dialogue history. We compare each model's reasoning with this rationale and categorize the coverage as complete, partial, or no coverage. Figure~\ref{fig:heatmap} reports the joint distribution between rationale coverage and predicted labels, while Figure~\ref{fig:coverage} summarizes the same results by label correctness and rationale coverage.

\section{UC-Bench}
\label{appendix:uc-bench}
\subsection{Construction Process}
\paragraph{WildChat-UC}
We first sampled 4K multi-turn dialogues from WildChat, with 2K in English and 2K in Chinese. We then cleaned the raw data in two stages. First, we applied rule-based filters to remove samples with abnormal dialogue structures, empty turns, missing fields, or obvious errors. Second, we used an LLM to filter out dialogues that were invalid, meaningless, or unsuitable as task-oriented dialogues at the semantic level. This process yielded 1,647 valid dialogues.

Next, we used an LLM to annotate the task category of each dialogue and selected task types suitable for analyzing multi-turn constraint relations. Specifically, we retained six categories: writing, roleplay, coding, text editing, summarization, and planning, resulting in 713 candidate dialogues. These tasks typically involve clear user intents and relatively stable task specifications, which makes it easier to determine whether a subsequent user utterance continues, revises, or conflicts with prior requirements. In contrast, although open-domain QA and casual chat account for a large proportion of the raw data, their interaction goals are often one-off factual queries or open-ended responses. They therefore exhibit weaker continuity of historical constraints and are less suitable as benchmark sources. We further used an LLM to score the 713 candidate samples on a 1--5 scale according to their suitability for implicit conflict analysis and sample construction, and retained the 152 samples with a score of 5 for the next stage.

\paragraph{Annotation.}
For each sample, two annotators first independently examined the relation between the final user utterance and the dialogue history. If the utterance could be directly identified as an implicit conflict or an explicit revision, it was assigned the corresponding label; otherwise, it was labeled as normal dialogue. We measured inter-annotator agreement on this three-way classification task using Cohen's $\kappa$, which reached 0.8696, indicating high agreement. Samples with inconsistent annotations were further adjudicated by a third annotator to determine the final label. The initial review showed that most instances were normal dialogue. Only a small number could be directly labeled as explicit revision, and no instances could be reliably identified as implicit conflict. These results indicate that natural sampling alone is insufficient to obtain enough conflict-related instances for systematic analysis.

To construct a benchmark better suited for controlled analysis of this phenomenon, we further performed manual construction on samples labeled as normal dialogue. While keeping the original dialogue history unchanged, annotators added one subsequent user utterance that formed either an implicit conflict or an explicit revision with the existing context. Because this process is open-ended and has no unique gold answer, we did not compute agreement on the generated utterance text itself. Instead, we adopted a quality-control procedure based on independent rewriting by two annotators and review by three evaluators.

Specifically, two annotators first independently reviewed each original dialogue and determined whether it was more suitable to be rewritten as explicit revision, rewritten as implicit conflict, or kept as normal dialogue. After discussion, they agreed on the target category and then independently wrote a follow-up user utterance. Three evaluators then reviewed the two candidate utterances, focusing on whether the target category was valid and the expression was natural. A candidate utterance was accepted if it was approved by at least two evaluators. If both candidates were accepted, the three evaluators further voted to select the better version, and the candidate with the most votes was retained. If only one candidate was accepted, it was retained directly. If neither candidate was accepted, the two annotators revised the sample until one candidate passed the review. For this review process, we report the three-way exact agreement rate and average pairwise agreement as descriptive agreement metrics. Across all candidate utterances, the three evaluators achieved a three-way exact agreement rate of 88.82\% and an average pairwise agreement of 92.54\%, indicating high consistency in the review process.

\paragraph{Dataset Composition.}
To support a focused analysis of implicit conflict, we set the ratio of implicit conflict, explicit revision, and normal dialogue to 3:1:1. We also maintained an approximately balanced distribution of Chinese and English samples to reduce potential confounding effects from language differences.

Following these criteria, the final subset contains 100 samples: 60 implicit conflict, 20 explicit revision, and 20 normal dialogue samples, including 51 Chinese samples and 49 English samples. This composition was constrained by the available high-quality sample pool. Since the pool contained 30 English implicit conflict instances, we selected 30 Chinese implicit conflict instances to balance this key class across the two languages. We then added explicit revision and normal dialogue samples to match the predefined 3:1:1 class ratio.

Following the construction of WildChat-UC, we set the target size of both ShareGPT-UC and LMSYS-UC to 100 samples and aimed to keep the three classes, implicit conflict, explicit revision, and normal dialogue, approximately balanced at a 1:1:1 ratio.

\paragraph{ShareGPT-UC}
For ShareGPT-UC, we applied the same rule-based filtering and LLM-based semantic filtering procedure as in WildChat-UC. This yielded 1,000 candidate samples, with 2,117 samples filtered out. Because Chinese data accounted for only a small portion of the candidates, we retained all 113 Chinese samples and supplemented them with 87 English samples, resulting in 200 samples for manual annotation and rewriting. We then followed the same annotation and quality-control protocols used for WildChat-UC. The inter-annotator agreement on the three-way classification task reached a Cohen's $\kappa$ of 0.7615. The three evaluators achieved a three-way exact agreement rate of 87.50\% and an average pairwise agreement of 91.67\% during review. After rewriting and review, the retained valid samples included 108 normal dialogue, 42 explicit revision, and 40 implicit conflict samples. To match the target size and approximate the 1:1:1 class ratio, we selected 33 normal dialogue, 33 explicit revision, and 34 implicit conflict samples to form ShareGPT-UC.

\paragraph{LMSYS-UC}
For LMSYS-UC, we used only English data because the Chinese data in LMSYS was of relatively low quality. We applied the same rule-based filtering and LLM-based semantic filtering procedure as in WildChat-UC, obtaining 200 English candidate samples after filtering out 1,562 samples. These candidates were then annotated and rewritten following the same annotation and quality-control protocols used for WildChat-UC. The inter-annotator agreement on the three-way classification task reached a Cohen's $\kappa$ of 0.6929. The three evaluators achieved a three-way exact agreement rate of 86.25\% and an average pairwise agreement of 90.83\% during review. After rewriting and review, the retained samples consisted of 134 normal dialogue, 33 explicit revision, and 33 implicit conflict samples. To match the target size and approximate the 1:1:1 class ratio, we selected 34 normal dialogue, 33 explicit revision, and 33 implicit conflict samples to form LMSYS-UC.
\subsection{Labeling Standards}

To ensure consistency in the rewriting process, annotators were required to first briefly analyze the original dialogue before writing the follow-up user utterance. Annotators were allowed to use search engines or LLMs to help understand the dialogue background, domain-specific terms, or task content. However, the final category decision and rewritten utterance had to be verified by the annotators themselves. Each sample was recorded using the following format.

\begin{enumerate}
\item \textbf{Task Goal.}

Summarize, in one sentence, the core task that the user expects the model to complete in the dialogue.

\textit{Requirements:}
\begin{itemize}
    \item Summarize only the task itself without expanding into details.
    \item Do not introduce a new task that is absent from the original dialogue.
    \item Keep the description concise and clear.
\end{itemize}

\item \textbf{Target Category.}

Specify the target category to be constructed for the follow-up user utterance:
\begin{itemize}
    \item Implicit Conflict
    \item Explicit Revision
\end{itemize}

\item \textbf{Rewriting Rationale.}

Explain the core basis on which the rewritten utterance supports the target category.

If the target category is implicit conflict, the annotator should specify which prior user requirement becomes implicitly incompatible with the new utterance. This incompatibility should not rely on explicit negation or direct modification, but should instead depend on the dialogue context, task goal, object attributes, stylistic requirements, or implicit constraints.

If the target category is explicit revision, the annotator should specify which prior user requirement is explicitly modified, replaced, or canceled by the new utterance. The modification should contain clear revision signals in the text.

\item \textbf{Follow-up User Utterance.}

Write a new follow-up user utterance while keeping the original dialogue history unchanged.

\textit{Requirements:}
\begin{itemize}
    \item Preserve the original task goal.
    \item Ensure that the utterance is natural and consistent with how real users continue to express requirements.
    \item Introduce only the minimal semantic change needed for the target category.
    \item Do not add a new task or new background information unrelated to the original task.
    \item Do not explicitly explain the conflict or reveal the target category.
    \item For implicit conflict, avoid obvious negation, replacement, cancellation, or rewriting signals.
    \item For explicit revision, include a clear revision intent so that it can be distinguished from implicit conflict.
\end{itemize}

\end{enumerate}

\section{Additional Method Details}
\subsection{Task Eligibility Filter}
\label{appendix:task_eligibility_filter}
Not all multi-turn dialogues are suitable for user-side conflict synthesis. Open-domain QA, casual conversation, and dialogues with unstable goals often lack clear task boundaries or persistent constraints, making it difficult to construct interpretable conflict samples. SynUC therefore uses an LLM to assess the task eligibility of each seed dialogue. We retain dialogues with well-defined task goals and constraint structures, including writing, roleplay, coding, text editing, summarization, and planning.
\par
Dialogues that fail the eligibility check are used only as candidates for normal dialogue, whereas eligible dialogues are routed to the explicit revision and implicit conflict synthesis pipelines. To preserve task diversity among normal dialogue samples, SynUC additionally assigns each eligible dialogue to the pipeline with probability $\alpha$. We set $\alpha = 10\%$ by default.

\subsection{SynUC Algorithm}
The pseudocode for SynUC is shown in Algorithm~\ref{alg:synuc}.
\algrenewcommand\algorithmiccomment[1]{\ {\color{gray}\# #1}}
\begin{algorithm*}[t]
\caption{SynUC Data Synthesis}
\label{alg:synuc}
\small
\begin{algorithmic}[1]
\Require Seed dialogues $\mathcal{D}$, SPEAKING schemas $\mathcal{S}$,
repositories $\{\mathcal{R}_{s_k}\}_{k=1}^{M}$,
buffers $\{\mathcal{B}_{s_k}\}_{k=1}^{M}$,
confidence threshold $\tau$
\Ensure Accepted samples $\mathcal{A}$ and discarded samples $\mathcal{Z}$

\State $\mathcal{A}\gets\emptyset,\quad \mathcal{Z}\gets\emptyset$

\ForAll{$D^{(t)}\in\mathcal{D}$}
    \State $\mathcal{Y}\gets\{\mathrm{ER},\mathrm{IC}\}$ if
    $\Call{TaskFilter}{D^{(t)}}=1$, otherwise $\{\mathrm{ND}\}$
    \Comment{route seed dialogues}

    \ForAll{$y\in\mathcal{Y}$}
        \If{$y=\mathrm{ND}$}
            \State Split $D^{(t)}$ into history $D^{(t-1)}$
            and candidate utterance $u_t$
            \Comment{reuse the original follow-up}
            \State $D_{\mathrm{h}}\gets D^{(t-1)}$
            \State $\langle\mathcal{T},\mathcal{C}\rangle
            \gets\Call{ExtractConstraints}{D_{\mathrm{h}}}$
            \State $u\gets u_t,\quad \pi\gets\emptyset$
        \Else
            \State $D_{\mathrm{h}}\gets D^{(t)}$
            \State $\langle\mathcal{T},\mathcal{C}\rangle
            \gets\Call{ExtractConstraints}{D_{\mathrm{h}}}$

            \If{$y=\mathrm{IC}$}
                \State $\mathcal{C}_y\gets\mathcal{C}$
                \Comment{IC may involve explicit or implicit constraints}
            \Else
                \State $\mathcal{C}_y\gets
                \{c\in\mathcal{C}\mid\tau(c)=\mathrm{explicit}\}$
                \Comment{ER only revises explicit constraints}
            \EndIf

            \State $(c_h,s_k)\gets
            \Call{MatchSchema}{\mathcal{C}_y,\mathcal{S}}$
            \Comment{select anchor constraint and schema}

            \If{$y=\mathrm{IC}$}
                \State $c_{t+1}\gets
                \Call{GenerateConstraint}{c_h,s_k,\mathcal{R}_{s_k}}$
                \Comment{use trajectory memory}
                \State $u\gets
                \Call{RealizeAsContinuation}{D_{\mathrm{h}},c_{t+1}}$
                \Comment{avoid explicit revision signals}
            \Else
                \State $c_{t+1}\gets
                \Call{GenerateConstraint}{c_h,s_k}$
                \State $u\gets
                \Call{RealizeAsRevision}{D_{\mathrm{h}},c_{t+1}}$
                \Comment{express an explicit update}
            \EndIf

            \State $\pi\gets
            (D_{\mathrm{h}},\mathcal{C},c_h,c_{t+1},u,s_k)$
            \Comment{synthesis trajectory}
        \EndIf

        \State $v_{\mathrm{a}}\gets 1$ if $y=\mathrm{ND}$,
        otherwise
        $\mathbb{I}\!\left[\mathcal{J}_{y}
        (c_h,c_{t+1},u)=1\right]$
        \Comment{anchor-level verification}

        \State $\hat{\mathcal{C}}\gets
        \Call{ExtractNewConstraints}{u}$

        \State $(v_{\mathrm{g}},q)\gets
        \Call{GlobalVerify}
        {D_{\mathrm{h}},\mathcal{C},\hat{\mathcal{C}},u,y}$
        \Comment{global verification and confidence scoring}

        \State $v\gets
        v_{\mathrm{a}}\land v_{\mathrm{g}}
        \land\mathbb{I}[q\ge\tau]$
        \Comment{confidence filtering}

        \If{$v=1$}
            \State $\mathcal{A}\gets
            \mathcal{A}\cup\{(D_{\mathrm{h}},u,y)\}$
        \Else
            \State $\mathcal{Z}\gets
            \mathcal{Z}\cup\{(D_{\mathrm{h}},u,y)\}$
        \EndIf

        \If{$y=\mathrm{IC}$}
            \State $\Call{WriteBuffer}{\mathcal{B}_{s_k},\pi,v}$
            \Comment{store trajectory feedback}

            \If{$\mathcal{B}_{s_k}$ reaches the update threshold}
                \State $\mathcal{R}_{s_k}\gets
                \Call{UpdateRepository}
                {\mathcal{R}_{s_k},\mathcal{B}_{s_k}}$
                \Comment{delayed memory update}
                \State $\mathcal{B}_{s_k}\gets\emptyset$
            \EndIf
        \EndIf
    \EndFor
\EndFor

\State \Return $\mathcal{A},\mathcal{Z}$
\end{algorithmic}
\end{algorithm*}

\subsection{The SPEAKING Framework}
\label{appendix:speaking}
SynUC adopts Hymes's SPEAKING framework as a structured prior for organizing communicative constraints. The framework decomposes a communicative event into eight dimensions that capture the contextual factors underlying utterance production and interpretation. Table~\ref{tab:speaking_dimensions} summarizes the meaning of each dimension and provides examples of constraint changes within a task.

\begin{table*}[t]
\caption{The SPEAKING dimensions and examples of constraint changes.}
\label{tab:speaking_dimensions}
\centering
\small
\renewcommand{\arraystretch}{1.3}
\begin{tabularx}{\textwidth}{
>{\raggedright\arraybackslash}p{0.13\textwidth}
>{\raggedright\arraybackslash}p{0.38\textwidth}
>{\raggedright\arraybackslash}X}
\toprule
\textbf{Dimension} & \textbf{Definition} & \textbf{Example} \\
\midrule
Setting and Scene &
Setting refers to the time, place, and physical environment in which a communicative event occurs. Scene refers to the psychological setting or culturally recognizable situation associated with the event. &
Task: writing an event introduction. Historical constraint: the introduction is intended for an offline campus presentation. New constraint: the introduction should be suitable for the opening of an online livestream. \\

Participants &
Participants refer to the people or groups involved in a communicative event, including speakers, hearers, audiences, addressees, and other relevant roles. &
Task: rewriting a science explanation. Historical constraint: the explanation targets middle school students. New constraint: the explanation targets domain experts. \\

Ends &
Ends refer to the purposes, goals, and expected outcomes of a communicative event, including the intentions or effects that participants aim to achieve through communication. &
Task: writing product copy. Historical constraint: the copy aims to encourage users to purchase the product. New constraint: the copy aims to remind users to carefully assess whether the product is suitable for them. \\

Act Sequence &
Act Sequence refers to the organization of verbal form and content in a communicative event, including how speech acts unfold and are ordered. &
Task: organizing a response. Historical constraint: present the conclusion first, followed by the rationale. New constraint: introduce the background and limitations first, and present the final conclusion at the end. \\

Key &
Key refers to the tone, manner, or overall register of a communicative event, such as formal, casual, serious, humorous, or sarcastic. &
Task: polishing a notice. Historical constraint: keep the tone light and friendly. New constraint: use a formal and serious tone. \\

Instrumentalities &
Instrumentalities refer to the channels, codes, languages, registers, or media used for communication, such as spoken language, written language, a specific language, dialect, or digital medium. &
Task: rewriting promotional content. Historical constraint: use formal written Chinese. New constraint: use English expressions suitable for social media. \\

Norms &
Norms refer to the social and cultural rules that govern interaction and interpretation, including participation conventions, politeness norms, turn-taking rules, and interpretive conventions. &
Task: writing feedback. Historical constraint: the feedback should be tactful and avoid directly negating the recipient. New constraint: the feedback should directly identify the problems and state clear revision requirements. \\

Genre &
Genre refers to culturally recognizable types of communicative events or discourse forms, such as lectures, stories, interviews, letters, reports, and reviews. &
Task: presenting project results. Historical constraint: write the content as a formal research report. New constraint: write the content as an interview-style Q\&A. \\
\bottomrule
\end{tabularx}
\end{table*}

\section{Additional Details of Pairwise Evaluation}
\label{sec:appendix-pairwise}
\paragraph{LLM-as-a-Judge Evaluation.} We use DeepSeek-V4-Flash as the judge model. For each implicit conflict instance, the judge receives the human-annotated conflict rationale and the reasoning outputs of the two models under comparison. It evaluates whether each reasoning output accurately identifies the historical and new constraints that give rise to the conflict. A comparison is labeled \textit{Win} if the reasoning produced by SynUC (DR-GRPO$^\dagger$) aligns more closely with the human-annotated conflict rationale, \textit{Tie} if the two outputs are comparable in quality, and \textit{Loss} if the baseline output is superior. To reduce potential positional bias, we randomly vary the presentation order of the two outputs.

\paragraph{Human Evaluation.} Two evaluators independently assess each pair of outputs using the same criteria as the LLM judge. Each comparison is assigned one of three labels: \textit{Win}, \textit{Tie}, or \textit{Loss}. After independent annotation, we measure inter-annotator agreement using raw agreement and Cohen's $\kappa$. For instances on which the two evaluators disagree, a third evaluator independently adjudicates the case, and the third evaluator's judgment is used as the final human label.

\begin{figure*}
    \centering
    \includegraphics[width=0.9\linewidth]{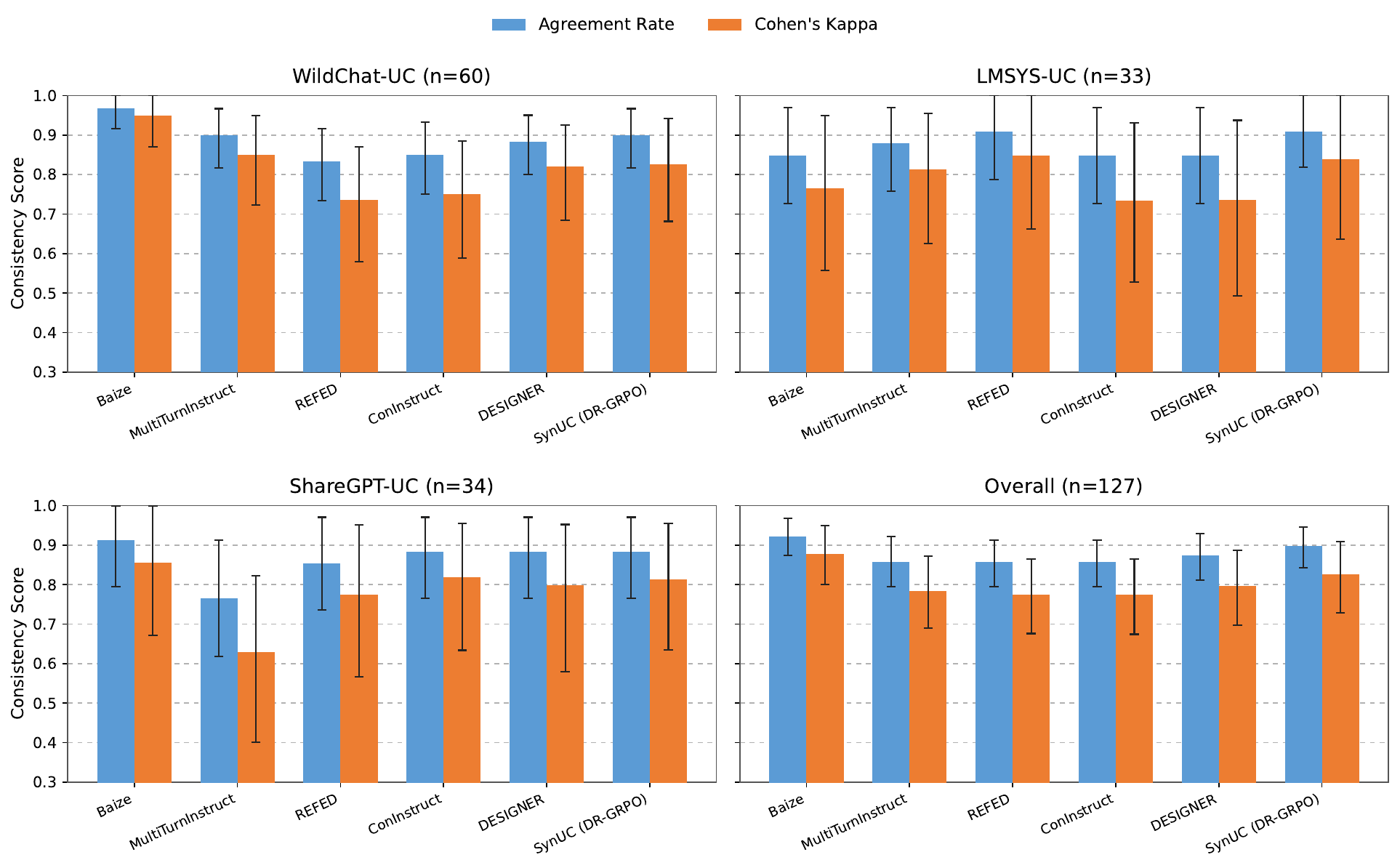}
    \caption{Inter-annotator agreement between two human evaluators. Raw agreement and Cohen's $\kappa$ are reported across different UC-Bench subsets, with error bars indicating 95\% confidence intervals.}
    \label{fig:annotator_agreement}
\end{figure*}

\begin{figure*}
    \centering
    \includegraphics[width=0.9\linewidth]{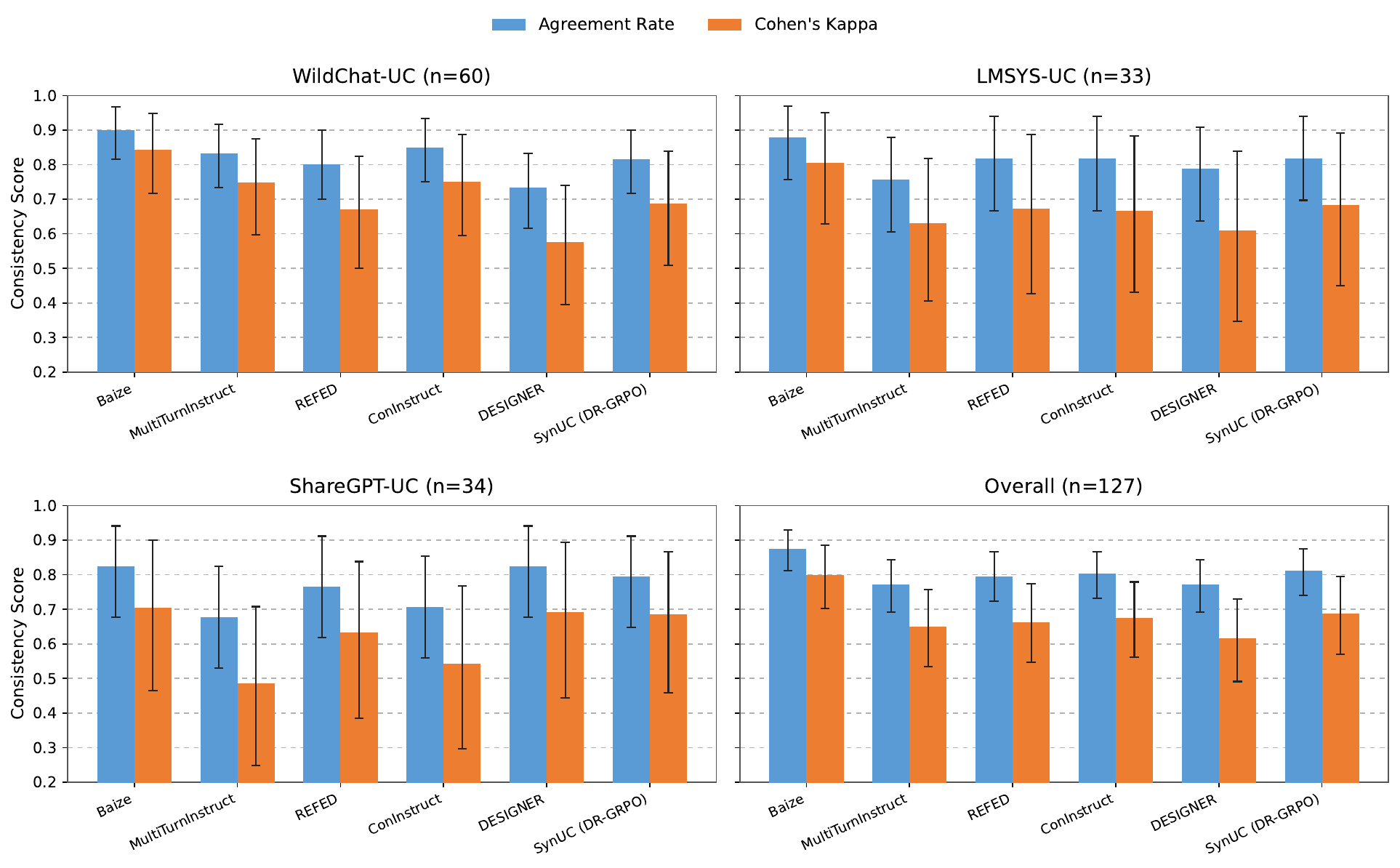}
    \caption{Consistency between LLM-based judgments and final human evaluations. Raw agreement and Cohen's $\kappa$ are reported on different UC-Bench subsets, with error bars indicating 95\% confidence intervals.}
    \label{fig:human_llm_agreement}
\end{figure*}

To evaluate the reliability of the annotation process, we report the raw agreement and Cohen's $\kappa$ coefficient between the two human evaluators, as shown in Figure~\ref{fig:annotator_agreement}. Furthermore, we assess the consistency between the LLM-based judgments and the final human judgments using the same metrics, as shown in Figure~\ref{fig:human_llm_agreement}.

\section{Case Study}
Table~\ref{tab:case_study} presents the reasoning processes of Qwen3.5-4B before and after fine-tuning on UC-Data under the same dialogue context. For clarity, the original outputs are organized into a standardized format.

\begin{table*}[t]
\centering
\caption{Case study comparing the structured reasoning of Qwen3.5-4B before and after fine-tuning on UC-Data, with the verbose output before fine-tuning condensed for clarity.}
\label{tab:case_study}
\footnotesize
\setlength{\tabcolsep}{6pt}
\renewcommand{\arraystretch}{1.12}

\begin{tabularx}{\textwidth}{
    >{\raggedright\arraybackslash}p{3.1cm}
    >{\raggedright\arraybackslash}X
}
\toprule
\textbf{Model Setting} &
\textbf{Structured Reasoning Output} \\
\midrule

\multicolumn{2}{
    >{\raggedright\arraybackslash}
    p{\dimexpr\textwidth-2\tabcolsep\relax}
}{
\textbf{[Dialogue History]}

\textbf{User:} I want to plan a short weekend trip for my wife and kids.
Could you help me make a relaxing itinerary?

\textbf{Assistant:} Sure. I can help you design a family-friendly weekend
itinerary with a relaxed pace.

\textbf{User:} Please avoid crowded tourist attractions. We prefer quiet
places where the kids can also enjoy themselves.

\textbf{Assistant:} Got it. I will focus on less crowded, family-friendly
places and avoid busy tourist spots.

\medskip
\textbf{[Subsequent Query]}

\textbf{User:} Great. Can you also add a wine-pairing dinner and a
late-night jazz show to the itinerary?
} \\

\midrule

\textbf{Qwen3.5-4B}
\newline
\textit{Before Fine-Tuning}
&
\textbf{[Task Level]}

The user continues the existing travel-planning task by adding specific
activities to the itinerary.

\medskip
\textbf{[Historical Constraints]}

The trip should be short and relaxing, accommodate the user's wife and
children, avoid crowded tourist attractions, and focus on quiet places that
the children can enjoy.

\medskip
\textbf{[New Constraints]}

The itinerary should include a wine-pairing dinner and a late-night jazz show.

\medskip
\textbf{[Constraint Compatibility Analysis]}

The newly requested activities may be less suitable for children. However,
the phrase ``also add'' is interpreted as a natural supplementation of the
existing itinerary rather than a conflict with the previous requirements.
Because the user does not explicitly reject the family-oriented constraints
and the activities may still be accommodated through suitable arrangements,
the request is treated as a refinement of the original plan.

\medskip
\textbf{[Final Judgment]}

Normal Dialogue
\\

\midrule

\textbf{Qwen3.5-4B}
\newline
\textit{Fine-Tuned on UC-Data}
&
\textbf{[Task Level]}

First, I extract the completed main tasks from the dialogue history: None

I also identify the current ongoing main task: Plan a short weekend trip for family

Next, I identify the subtasks already completed in the history: Create initial relaxing itinerary

The current subtask being handled is: Revise itinerary to avoid crowded attractions

Then I determine the task level continued by the subsequent query: \texttt{current\_subtask}

\medskip
\textbf{[Historical Constraints]}

At this continued task level, the historical constraints that need to be preserved are:
Explicit Constraint: \texttt{'content\_scope: avoid crowded tourist attractions'};
Explicit Constraint: \texttt{'content\_scope: quiet places where kids can enjoy'};
Implicit Constraint: infer \texttt{'content\_scope: family-friendly activities'}
from \texttt{'kids can also enjoy themselves'}

\medskip
\textbf{[New Constraints]}

Looking at the subsequent query itself, its explicit constraints are:
Explicit Constraint: \texttt{'content\_scope: add wine-pairing dinner'};
Explicit Constraint: \texttt{'content\_scope: add late-night jazz show'}

The constraints that can be implicitly inferred from it are:
Implicit Constraint: infer
\texttt{'audience: adults-only or couples-oriented activities'}
from \texttt{'wine-pairing dinner and late-night jazz show'}

\medskip
\textbf{[Constraint Compatibility Analysis]}

Based on this information, my judgment process is:
The subsequent query asks to add a wine-pairing dinner and a late-night jazz show.
These activities are typically oriented towards adults, not families with kids.
The historical implicit constraint requires family-friendly activities.
The new implicit constraint (adults-oriented) conflicts with the historical
implicit constraint (family-friendly).
This is an \texttt{H\_I\_N\_I} conflict: historical implicit constraint
(family-friendly) conflicts with new implicit constraint (adults-oriented).
The claimed new constraint matches the inferred implicit constraint.
Therefore, the query constitutes an Implicit Conflict.

\medskip
\textbf{[Final Judgment]}

Therefore, the final classification is: Implicit Conflict

\\

\bottomrule
\end{tabularx}
\end{table*}

\end{document}